\documentclass[11pt]{article}

\usepackage{acl}
\usepackage{placeins}

\usepackage[most]{tcolorbox}
\usepackage{times}
\usepackage{latexsym}
\fontsize{7}{8}\selectfont
\renewcommand{\arraystretch}{0.82}
\usepackage[most]{tcolorbox}   
\usepackage{xcolor}           
\usepackage{float}

\usepackage[table]{xcolor}
\usepackage{graphicx}
\usepackage{adjustbox}
\usepackage{multirow}
\usepackage{colortbl}

\usepackage[table]{xcolor}
\usepackage{placeins}

\usepackage{lmodern}

\usepackage{anyfontsize}

\usepackage[most]{tcolorbox}
\usepackage[most]{tcolorbox}
\usepackage{xcolor}
\definecolor{correctbg}{RGB}{220,245,220}
\definecolor{errorbg}{RGB}{245,220,220}

\newcommand{\labeltag}[2]{%
\tcbox[
on line,
boxrule=0pt,
arc=2pt,
left=3pt,
right=3pt,
top=1pt,
bottom=1pt,
colback=#1
]{#2}
}

\newtcolorbox{casebox}[1]{
colback=white,
colframe=black!40,
boxrule=0.6pt,
arc=2pt,
title=#1,
fonttitle=\bfseries\small,
coltitle=black,
boxrule=0.6pt,
colbacktitle=black!12,
arc=2pt,
left=4pt,
right=4pt,
top=4pt,
bottom=4pt
}

\definecolor{lightgraybg}{RGB}{245,245,245}
\definecolor{grayborder}{RGB}{170,170,170}
\definecolor{graytitle}{RGB}{120,120,120}

\tcbset{
mypanel/.style={
    enhanced,
    colback=lightgraybg,
    colframe=grayborder,
    coltitle=white,
    colbacktitle=graytitle,
    fonttitle=\bfseries,
    boxrule=0.6pt,
    arc=2pt,
    left=6pt,
    right=6pt,
    top=6pt,
    bottom=6pt,
    before skip=3pt,
    after skip=3pt
},
innerpanel/.style={
    enhanced,
    colback=white,
    colframe=grayborder,
    boxrule=0.4pt,
    arc=1.5pt,
    left=4pt,
    right=4pt,
    top=4pt,
    bottom=4pt,
    before skip=3pt,
    after skip=3pt
}
}

\definecolor{pos1}{RGB}{247,247,247}
\definecolor{pos2}{RGB}{240,240,240}
\definecolor{pos3}{RGB}{232,232,232}
\definecolor{pos4}{RGB}{223,223,223}
\definecolor{pos5}{RGB}{214,214,214}
\definecolor{pos6}{RGB}{204,204,204}
\definecolor{pos7}{RGB}{192,192,192}
\definecolor{pos8}{RGB}{178,178,178}
\definecolor{pos9}{RGB}{164,164,164}
\definecolor{pos10}{RGB}{157,157,157}

\definecolor{neg1}{RGB}{255,247,247}
\definecolor{neg2}{RGB}{255,240,240}
\definecolor{neg3}{RGB}{255,232,232}
\definecolor{neg4}{RGB}{250,223,223}
\definecolor{neg5}{RGB}{245,214,214}
\definecolor{neg6}{RGB}{240,204,204}
\definecolor{neg7}{RGB}{234,192,192}
\definecolor{neg8}{RGB}{226,178,178}
\definecolor{neg9}{RGB}{216,164,164}
\definecolor{neg10}{RGB}{205,150,150}

\definecolor{TitleBlue}{RGB}{0,0,139}   
\definecolor{PanelBack}{RGB}{245,246,255}

\usepackage[T1]{fontenc}

\usepackage[utf8]{inputenc}

\usepackage{microtype}

\usepackage{inconsolata}

\usepackage{graphicx}
\usepackage{dblfloatfix}
\usepackage{booktabs}   
\usepackage{multirow}   

\title{MoPrO-SD: Modular Prompt Optimization for Stance Detection}

\author{
\begin{tabular}{ccc}
Meysam Sabbaghan\textsuperscript{1,*} &
Nikesh Gyawali\textsuperscript{1,*} &
Arman Zareian Jahromi\textsuperscript{1} \\[5pt]
Sebastian Loera-Lozano\textsuperscript{1} &
Doina Caragea\textsuperscript{1} &
Cornelia Caragea\textsuperscript{2}
\end{tabular}
\\[8pt]
\textsuperscript{1}Kansas State University\quad             
\textsuperscript{2}University of Illinois Chicago \\[4pt]
\texttt{\{meysamsabbaghan,gnikesh,zareian,sloera,dcaragea\}@ksu.edu} \\
\texttt{cornelia@uic.edu}
}

\begin{document}
\maketitle
\begingroup
\renewcommand{\thefootnote}{*}
\footnotetext{Equal contribution.}
\endgroup
\begin{abstract}
Stance detection remains challenging for large language models (LLMs)
due to implicit opinions, sarcasm, and context-dependent expressions of
stance. Although automatic prompt optimization offers an alternative to
fine-tuning without updating model parameters, existing structured
approaches are not specifically designed around stance-label semantics
and class-specific ambiguities. We introduce
\textbf{MoPrO-SD}, a stance-specific prompt optimization framework that
decomposes prompts into functional semantic modules and refines them
sequentially using LLM-based feedback. We study three refinement
strategies---instruction-only optimization (\textbf{I}), instruction
with reasoning (\textbf{I+R}), and instruction with per-example
reasoning (\textbf{I+R+E})---using class-conditioned errors for
stance-label refinement. We also introduce \textbf{VaccineStance}, a
new human-annotated benchmark for vaccine stance detection. Experiments
across four datasets and five open-weight LLMs show that MoPrO-SD
improves upon the human-written Base prompt in many settings.
On SemEval-2016, class-conditioned feedback achieves higher mean
Macro-F1 than pooled feedback for four of five backbones.
Both reasoning-based strategies, I+R and I+R+E, achieve strong
performance, with their relative effectiveness varying across models
and datasets, while remaining competitive with recent prompt
optimization methods.
The MoPrO-SD code and VaccineStance dataset are available on GitHub.\footnote{
\url{https://anonymous.4open.science/r/labeled-vaccines-tweets}}
\end{abstract}

\section{Introduction}

Stance detection identifies the attitude toward a target or claim, typically categorized as \textit{In-Favor}, \textit{Against}, or \textit{Neutral}~\citep{mohammad2016semeval}. It supports applications including misinformation detection, political opinion analysis, public health monitoring, and social media analytics~\citep{kuccuk2020stance,aldayel2021stance}. Despite advances in large language models (LLMs), stance detection remains challenging due to sarcasm, implicit opinions, and context-dependent expressions~\citep{yakura2024evaluating,gole2024sarcasm,li2025sarcasm}.

Prompting offers an alternative to fine-tuning without updating model parameters, but LLMs are sensitive to prompt wording, demonstrations, and example ordering~\citep{zhao2021calibrate,lu2022fantastically,shi2024larger}. Semantically similar prompts can yield different performance~\citep{kojima2022large,zhou2022large}, and effective prompts are often task- and model-dependent~\citep{chen2023you}, motivating automated prompt optimization~\citep{chang2024efficient}.

Recent methods include OPRO~\citep{yang2024large}, EvoPrompt~\citep{guo2024evoprompt}, and CAPO~\citep{pmlr-v293-zehle25a}, while UniPrompt~\citep{juneja-etal-2025-task} and SCULPT~\citep{kumar-etal-2025-sculpt} optimize structured prompt components. Reasoning-based prompting has also shown promise for stance detection~\citep{ma2024chain,gul2024stance,aiyappa2024benchmarking}. However, existing optimization frameworks are largely task-general rather than organized around stance-specific class structure, where guidelines, label definitions, and output constraints serve distinct roles.

We introduce \textbf{MoPrO-SD} (\textbf{Mo}dular \textbf{Pr}ompt \textbf{O}ptimization for \textbf{S}tance \textbf{D}etection), which decomposes a stance prompt into six functional modules and refines them sequentially. We study instruction-only (\textbf{I}), instruction with reasoning (\textbf{I+R}), and instruction with per-example reasoning (\textbf{I+R+E}) optimization. Reasoning-based strategies use misclassification feedback for targeted refinement, with class-conditioned errors informing stance-label modules.

We evaluate MoPrO-SD on \textit{SemEval-2016}~\citep{mohammad2016semeval}, \textit{COVID-19-Stance}~\citep{glandt2021stance}, \textit{GunStance}~\citep{gyawali2024gunstance}, and \textit{VaccineStance} using five open-weight LLMs, comparing against OPRO~\citep{yang2024large}, CAPO~\citep{pmlr-v293-zehle25a}, EvoPrompt-GA/DE~\citep{guo2024evoprompt}, UniPrompt~\citep{juneja-etal-2025-task}, and SCULPT~\citep{kumar-etal-2025-sculpt}. MoPrO-SD improves over the human-written base prompt across many model--dataset settings. We also introduce \textbf{VaccineStance}, a new human-annotated benchmark constructed from vaccine-related social media posts collected between 2013 and 2022.

Our contributions are:
\begin{itemize}
    \item We introduce \textbf{MoPrO-SD}, combining stance-specific modular refinement with class-conditioned error feedback.
    \item We evaluate it across four datasets and five LLMs against established baselines, with analyses of modularity, optimization budget, cost, module contributions, few-shot prompting, and external-reasoner transfer.
    \item We introduce \textbf{VaccineStance}, a new human-annotated benchmark.
\end{itemize}

\section{Related Work}

\paragraph{Prompting for Stance Detection.}
LLM-based stance detection uses prompting and
reasoning strategies~\citep{pangtey2025large}. Chain-of-Stance
\citep{ma2024chain} and Stance Reasoner
\citep{taranukhin-etal-2024-stance} use explicit reasoning, while other
work explores rationale-augmented prompting~\citep{lee2024logic} and
zero-shot instruction and decoding strategies~\citep{aiyappa2024benchmarking}.
Supervised adaptation~\citep{gul2024stance} can achieve strong performance
but requires additional training. In contrast, we optimize the stance
prompt without updating the target LLM.

\paragraph{Automated and Structured Prompt Optimization.}
Automatic prompt optimization spans instruction generation, feedback,
search, reinforcement learning, and evolutionary approaches
\citep{ramnath2025systematic}. Representative methods include APE
\citep{zhou2022large}, ProTeGi~\citep{pryzant2023automatic}, RLPrompt
\citep{deng2022rlprompt}, MAPO~\citep{chen-etal-2023-mapo}, PRewrite
\citep{kong2024prewrite}, Auto Evol-Instruct~\citep{zeng2024automatic},
PromptAgent~\citep{wang2024promptagent}, and Align-Pro
\citep{trivedi2025alignpro}. OPRO~\citep{yang2024large} uses evaluated
solutions to guide optimization, EvoPrompt~\citep{guo2024evoprompt}
combines LLMs with evolutionary algorithms, and CAPO
\citep{pmlr-v293-zehle25a} incorporates optimization cost.

Structured approaches include Decomposed Prompting
\citep{khot2023decomposed}, modular LLM programs in DSPy
\citep{khattab2023dspy}, structure-aware optimization in Symbolic Prompt
Program Search~\citep{schnabel-neville-2024-symbolic}, content--format
optimization in CFPO~\citep{liu2025beyond}, and section-wise modular
prompt optimization~\citep{sharma2026modular}. DS2
\citep{pang2025improving} and DeLLMa~\citep{liu2024dellma} explore
data- or decision-aware selection, while Prompt Selection and
Augmentation~\citep{wu2024prompt}, BRIDGE~\citep{wan2025few},
APT-Pipe~\citep{zhu2024apt}, and Mixture of Demonstrations
\citep{wang2024mixture} study demonstration selection and organization.

Related work includes UniPrompt~\citep{juneja-etal-2025-task},
SCULPT~\citep{kumar-etal-2025-sculpt}, and modular prompt
optimization~\citep{sharma2026modular}. While these methods optimize
structured prompt components, MoPrO-SD specializes this paradigm to
stance detection through fixed functional modules, sequential refinement,
and class-conditioned misclassification feedback for stance-label modules.
\begin{figure*}
  \centering
  \includegraphics[width=\linewidth]{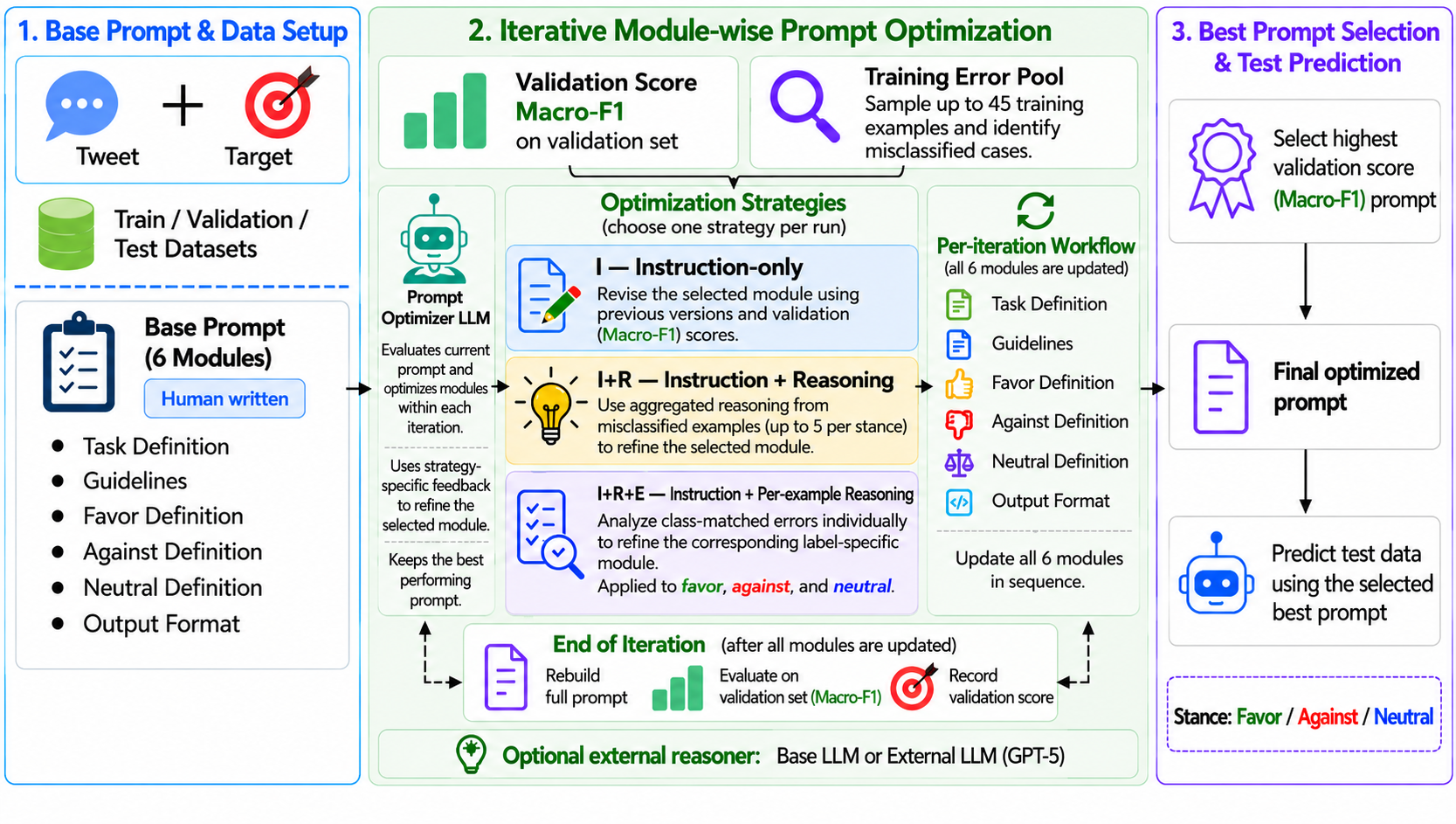}
  \caption{Overview of MoPrO-SD. Starting from a six-module base prompt, MoPrO-SD sequentially refines the modules using instruction-only (I), instruction with reasoning (I+R), or instruction with per-example reasoning (I+R+E). After each iteration, the updated prompt is evaluated using validation Macro-F1, and the highest-scoring prompt is selected for final test evaluation.}
  \label{fig:iterative_prompt_improvement}
\end{figure*}

\section{Methodology}

We introduce \textbf{MoPrO-SD}, a \textbf{Mo}dular \textbf{Pr}ompt
\textbf{O}ptimization framework for \textbf{S}tance \textbf{D}etection.
MoPrO-SD decomposes a stance prompt into semantically distinct modules
and iteratively refines them using LLM-based feedback. We study three
strategies: instruction-only (\textbf{I}), instruction with reasoning
(\textbf{I+R}), and instruction with per-example reasoning
(\textbf{I+R+E}). Figure~\ref{fig:iterative_prompt_improvement}
summarizes the framework.

\subsection{Stance Detection Prompt Modules}

The base prompt consists of six modules: \textit{task definition},
\textit{guidelines}, \textit{favor definition}, \textit{against
definition}, \textit{neutral definition}, and \textit{output format}.
This design follows prior work on decomposing long prompts for more
stable and controllable optimization~\citep{kumar-etal-2025-sculpt,
khot2023decomposed,tang2025unleashing}. The task module specifies the
stance detection objective; guidelines provide general decision criteria,
including sarcasm, implicit stance, and context-dependent expressions;
the three label-specific modules define \textit{favor}, \textit{against},
and \textit{neutral} and may include representative examples to reduce
label ambiguity~\citep{mohammad2016semeval}; and the output module
enforces label-only predictions. The complete modular base prompt is
provided in Appendix~\ref{sec:complete_prompt}.

\subsection{Proposed MoPrO-SD Framework}

Starting from a manually designed six-module base prompt, MoPrO-SD first
evaluates the full prompt on a validation set using Macro-F1 and then
refines the modules sequentially in a fixed order: \textit{task
definition}, \textit{guidelines}, \textit{favor definition},
\textit{against definition}, \textit{neutral definition}, and
\textit{output format}. During each update, only the target module is
rewritten using feedback determined by the optimization strategy below.

After all six modules are processed, the full prompt is re-evaluated on
the validation set. One pass over the six modules constitutes an
iteration, and the prompt with the highest validation Macro-F1 across
iterations is selected for final test evaluation. This procedure
resembles coordinate descent in prompt space, preserving prompt
substructures while refining individual modules. In the main experiments,
the optimizer/reasoner is the same backbone as the stance classifier; we
separately evaluate a stronger external optimizer.

\subsection{Prompt Optimization Strategies}

\paragraph{Instruction-only Optimization (I).}
All six modules are refined using instruction-only updates. For each
module, the optimizer uses its previous variants and the
validation Macro-F1 recorded for their corresponding iterations to
propose a new version, without explicit analysis of misclassified
examples~\citep{zhou2022large}.

\paragraph{Instruction with Reasoning Optimization (I+R).}
The \textit{guidelines} and three label-specific modules are refined
using aggregated reasoning over misclassified training examples, while
the \textit{task definition} and \textit{output format} use
instruction-only refinement. For each label-specific module, only errors
whose gold label corresponds to that module are used; for example, the
\textit{favor definition} uses errors with gold label \textit{favor}.
For the \textit{guidelines} module, errors from all three stance classes
are considered. The aggregated reasoning identifies problems in the
current module and guides its refinement~\citep{shin2025egot}.

\paragraph{Instruction with Per-example Reasoning Optimization (I+R+E).}
For the three label-specific modules, misclassified training examples
are analyzed individually. Following \citet{tang2025unleashing}, the
optimizer generates an explanation for each error, summarizes these
explanations, and uses the feedback to revise the corresponding
module. The \textit{task definition}, \textit{guidelines}, and
\textit{output format} continue to use instruction-only updates within
the same optimization loop. If no relevant errors are available for a
label-specific module, that module remains unchanged for that iteration.
Example meta-prompts are provided in
Appendix~\ref{sec:appendix_a_meta_prompts}.

I+R and I+R+E are treated as alternative configurations, as they differ
in both reasoning granularity and the treatment of the
\textit{guidelines} module.

\section{Experimental Setup}

\subsection{Stance Detection Datasets}

We evaluate MoPrO-SD on three existing stance detection benchmarks---
\textit{SemEval-2016 Task 6}, \textit{GunStance}, and
\textit{COVID-19-Stance}---and our new human-annotated
\textit{VaccineStance} benchmark.

\begin{enumerate}
    \item \textbf{SemEval-2016 Task 6}~\citep{mohammad2016semeval}
    contains tweets labeled as \textit{Favor}, \textit{Against}, or
    \textit{None} across multiple controversial targets. We exclude
    \textit{Donald Trump} because it has no training instances and map
    \textit{None} to \textit{Neutral} for consistency across datasets.

    \item \textbf{GunStance}~\citep{gyawali2024gunstance} contains
    stance-labeled tweets on gun-control debates, including the targets
    \textit{Banning Guns} and \textit{Regulating Guns}.

    \item \textbf{COVID-19-Stance}~\citep{glandt2021stance} contains
    stance-labeled tweets on COVID-19 measures, including face masks,
    school closures, and stay-at-home orders.

\item \textbf{VaccineStance} is a new human-annotated
stance detection benchmark introduced in this work.
We curated 2,000 vaccine-relevant tweets from the
corpus of Gyawali et al.~\cite{gyawali2025vaccine}
and obtained human annotations for \textit{Favor},
\textit{Against}, and \textit{Neutral}, with Fleiss'
$\kappa=0.778$. The dataset comprises 500 training,
500 validation, and 1,000 test instances.
\end{enumerate}

Detailed dataset statistics, VaccineStance construction and preprocessing,
and split distributions are provided in
Appendix~\ref{appendix_additional_details}.

\subsection{Large Language Models}

We evaluate five LLMs spanning different model families and scales:
\textbf{Phi-4-14B}~\citep{abdin2024phi},
\textbf{GPT-OSS-20B}~\citep{openai_gptoss_2025},
\textbf{Mistral-Small-3.2-24B}~\citep{mistral2025small31},
\textbf{Gemma-3-27B}~\citep{team2025gemma}, and
\textbf{Llama-3.3-70B}~\citep{meta_llama3p3_70b_2024}.
Tables abbreviate them as Phi, GPT-OSS, Mistral, Gemma, and Llama.

\subsection{Baselines}

We compare MoPrO-SD with the original zero-shot Base prompt and several
prompting baselines built on the same Base prompt: chain-of-thought
(Base+CoT)~\citep{wei2022chain}, few-shot prompting ($k=3$;
Base+FS)~\citep{brown2020language}, their combination (Base+CoT+FS),
and Chain-of-Stance (Base+CoS)~\citep{ma2024chain}.
We also compare against automated prompt optimization methods
OPRO~\citep{yang2024large}, CAPO~\citep{pmlr-v293-zehle25a}, and
EvoPrompt with differential evolution (DE) and genetic algorithm (GA)
variants~\citep{guo2024evoprompt}. Additionally, on SemEval-2016, we
compare with the structured prompt optimization methods
UniPrompt~\citep{juneja-etal-2025-task} and
SCULPT~\citep{kumar-etal-2025-sculpt} using Mistral, Gemma, and Llama.

\subsection{Implementation Details}

All experiments optimize prompts without fine-tuning model weights.
MoPrO-SD uses 10 optimization iterations, with the six modules processed
sequentially in a fixed order. After each iteration, the full
prompt is evaluated on the validation set, and the prompt with the
highest validation Macro-F1 across iterations is selected for final test
evaluation. All methods in Table~\ref{tab:aggregate_results} were run in
our experimental pipeline using fixed dataset splits and three
independent runs per method. Across runs, validation subsampling,
training-example sampling, and stochastic LLM generation can vary.
Optimization budgets vary across methods.

Unless otherwise specified, the same backbone serves as the stance
classifier and optimizer/reasoner. We additionally evaluate GPT-5 as a
stronger external reasoner~\citep{openai2025gpt5}. In frozen-module
ablations, one module remains fixed while the others are optimized.

\subsection{Evaluation Metrics}

We use three-class Macro-F1, computed as the unweighted mean of the
F1 scores for \textit{Favor}, \textit{Against}, and \textit{Neutral},
as the primary evaluation metric. This differs from the official
SemEval-2016 metric, which averages F1 over \textit{Favor} and
\textit{Against} only. Prompt selection is based on validation
Macro-F1, and final test results are reported as mean $\pm$ standard
deviation across three independent runs.
\begin{table*}[!ht]
\centering
\scriptsize
\setlength{\tabcolsep}{3.2pt}
\renewcommand{\arraystretch}{1.05}

\begin{tabular}{llccccc}
\toprule
\fontsize{8.7}{9.5}\selectfont
\textbf{Dataset} & \textbf{Method} & \textbf{Phi} & \textbf{GPT-OSS} &
\textbf{Mistral} & \textbf{Gemma} & \textbf{Llama} \\
\midrule

\multirow{12}{*}{SemEval}
& Base (zero-shot)
& 55.37 $\pm$ 0.12
& 60.02 $\pm$ 0.44
& 53.90 $\pm$ 0.28
& 65.82 $\pm$ 0.20
& 73.52 $\pm$ 0.14 \\

& Base+CoT
& \cellcolor{pos9}66.60 $\pm$ 0.74
& \cellcolor{neg2}57.90 $\pm$ 0.87
& \cellcolor{pos9}67.06 $\pm$ 0.25
& \cellcolor{pos3}67.45 $\pm$ 0.43
& \cellcolor{neg1}71.77 $\pm$ 0.30 \\

& Base+FS
& \cellcolor{pos3}57.90 $\pm$ 1.48
& \cellcolor{pos3}62.17 $\pm$ 0.79
& \cellcolor{pos8}64.69 $\pm$ 0.35
& \cellcolor{pos4}69.30 $\pm$ 0.05
& \cellcolor{neg2}72.28 $\pm$ 0.11 \\

& Base+CoT+FS
& \cellcolor{pos8}64.85 $\pm$ 0.44
& \cellcolor{pos1}59.89 $\pm$ 0.28
& \cellcolor{pos9}67.32 $\pm$ 0.56
& \cellcolor{pos3}66.98 $\pm$ 0.11
& \cellcolor{neg3}70.41 $\pm$ 0.12 \\

& Base+CoS
& \cellcolor{neg2}53.40 $\pm$ 0.81
& \cellcolor{pos1}60.18 $\pm$ 1.67
& \cellcolor{pos8}63.94 $\pm$ 0.37
& \cellcolor{neg6}59.92 $\pm$ 2.12
& \cellcolor{neg8}62.96 $\pm$ 1.45 \\

& OPRO
& \cellcolor{pos9}67.49 $\pm$ 0.74
& \cellcolor{pos6}66.59 $\pm$ 0.94
& \cellcolor{pos3}56.61 $\pm$ 3.19
& \cellcolor{neg2}64.76 $\pm$ 1.39
& \cellcolor{neg1}73.45 $\pm$ 0.27 \\

& CAPO
& \cellcolor{pos7}63.71 $\pm$ 0.60
& \cellcolor{pos7}\textbf{67.32 $\pm$ 1.05}
& \cellcolor{pos6}60.32 $\pm$ 0.12
& \cellcolor{pos2}67.80 $\pm$ 2.02
& \cellcolor{pos2}\textbf{74.87 $\pm$ 0.82} \\

& EvoPrompt(DE)
& \cellcolor{pos9}68.02 $\pm$ 0.37
& \cellcolor{pos6}65.59 $\pm$ 0.89
& \cellcolor{pos8}65.35 $\pm$ 1.47
& \cellcolor{pos3}67.96 $\pm$ 1.67
& \cellcolor{pos2}74.83 $\pm$ 1.14 \\

& EvoPrompt(GA)
& \cellcolor{pos8}67.28 $\pm$ 1.79
& \cellcolor{pos5}64.85 $\pm$ 1.53
& \cellcolor{pos7}62.32 $\pm$ 3.61
& \cellcolor{pos4}69.11 $\pm$ 0.39
& \cellcolor{neg2}72.51 $\pm$ 2.51 \\

& I (ours)
& \cellcolor{pos6}61.63 $\pm$ 1.38
& \cellcolor{neg1}59.62 $\pm$ 0.63
& \cellcolor{pos7}61.93 $\pm$ 4.00
& \cellcolor{pos3}68.26 $\pm$ 1.73
& \cellcolor{neg1}73.42 $\pm$ 0.43 \\

& I+R (ours)
& \cellcolor{pos9}\textbf{69.65 $\pm$ 1.62}
& \cellcolor{pos3}62.96 $\pm$ 1.26
& \cellcolor{pos9}67.85 $\pm$ 1.15
& \cellcolor{pos4}69.64 $\pm$ 2.08
& \cellcolor{pos2}74.85 $\pm$ 0.26 \\

& I+R+E (ours)
& \cellcolor{pos9}68.63 $\pm$ 0.53
& \cellcolor{pos6}66.10 $\pm$ 1.14
& \cellcolor{pos9}\textbf{71.12 $\pm$ 0.85}
& \cellcolor{pos4}\textbf{69.68 $\pm$ 1.14}
& \cellcolor{pos2}74.78 $\pm$ 0.38 \\

\midrule

\multirow{12}{*}{GunStance}
& Base (zero-shot)
& 39.55 $\pm$ 1.49
& 60.34 $\pm$ 0.77
& 56.24 $\pm$ 0.30
& 47.09 $\pm$ 0.14
& 54.85 $\pm$ 0.22 \\

& Base+CoT
& \cellcolor{pos10}\textbf{61.17 $\pm$ 0.26}
& \cellcolor{pos1}59.76 $\pm$ 0.56
& \cellcolor{pos3}57.55 $\pm$ 0.20
& \cellcolor{pos6}52.48 $\pm$ 0.09
& \cellcolor{pos3}56.27 $\pm$ 0.11 \\

& Base+FS
& \cellcolor{pos8}51.47 $\pm$ 1.54
& \cellcolor{pos1}60.60 $\pm$ 0.25
& \cellcolor{pos7}64.13 $\pm$ 0.17
& \cellcolor{pos8}58.03 $\pm$ 0.06
& \cellcolor{pos4}58.12 $\pm$ 0.20 \\

& Base+CoT+FS
& \cellcolor{pos10}59.32 $\pm$ 0.68
& \cellcolor{neg1}58.85 $\pm$ 0.64
& \cellcolor{pos6}61.72 $\pm$ 0.30
& \cellcolor{pos10}63.56 $\pm$ 0.38
& \cellcolor{pos8}59.01 $\pm$ 0.14 \\

& Base+CoS
& \cellcolor{pos4}43.16 $\pm$ 0.44
& \cellcolor{neg8}50.20 $\pm$ 1.25
& \cellcolor{pos6}62.55 $\pm$ 1.07
& \cellcolor{pos9}59.30 $\pm$ 1.59
& \cellcolor{neg6}49.74 $\pm$ 0.38 \\

& OPRO
& \cellcolor{pos10}55.90 $\pm$ 2.66
& \cellcolor{neg2}59.10 $\pm$ 1.80
& \cellcolor{pos2}57.36 $\pm$ 1.96
& \cellcolor{neg2}45.63 $\pm$ 0.45
& \cellcolor{neg1}54.28 $\pm$ 1.00 \\

& CAPO
& \cellcolor{pos10}56.69 $\pm$ 1.44
& \cellcolor{pos1}60.47 $\pm$ 0.46
& \cellcolor{pos6}62.19 $\pm$ 1.13
& \cellcolor{pos6}53.97 $\pm$ 2.06
& \cellcolor{neg1}54.38 $\pm$ 1.80 \\

& EvoPrompt(DE)
& \cellcolor{pos10}55.41 $\pm$ 3.67
& \cellcolor{neg2}58.42 $\pm$ 1.02
& \cellcolor{pos4}59.40 $\pm$ 1.86
& \cellcolor{pos6}52.20 $\pm$ 2.07
& \cellcolor{pos3}57.05 $\pm$ 1.02 \\

& EvoPrompt(GA)
& \cellcolor{pos10}56.14 $\pm$ 1.78
& \cellcolor{neg1}59.54 $\pm$ 1.64
& \cellcolor{pos3}59.14 $\pm$ 0.92
& \cellcolor{pos7}54.88 $\pm$ 2.16
& \cellcolor{neg2}53.83 $\pm$ 1.70 \\

& I (ours)
& \cellcolor{pos9}53.69 $\pm$ 2.41
& \cellcolor{pos2}61.82 $\pm$ 0.66
& \cellcolor{pos6}61.43 $\pm$ 2.45
& \cellcolor{pos8}58.53 $\pm$ 1.60
& \cellcolor{pos3}57.55 $\pm$ 0.44 \\

& I+R (ours)
& \cellcolor{pos10}60.99 $\pm$ 3.13
& \cellcolor{pos1}61.07 $\pm$ 0.71
& \cellcolor{pos7}63.68 $\pm$ 1.87
& \cellcolor{pos10}\textbf{64.05 $\pm$ 0.85}
& \cellcolor{pos5}58.98 $\pm$ 1.40 \\

& I+R+E (ours)
& \cellcolor{pos10}59.74 $\pm$ 0.26
& \cellcolor{pos2}\textbf{62.12 $\pm$ 1.76}
& \cellcolor{pos7}\textbf{65.01 $\pm$ 2.18}
& \cellcolor{pos9}60.21 $\pm$ 1.84
& \cellcolor{pos5}\textbf{59.03 $\pm$ 3.58} \\

\midrule

\multirow{12}{*}{COVID-19-Stance}
& Base (zero-shot)
& 50.36 $\pm$ 1.71
& 73.20 $\pm$ 1.45
& 61.15 $\pm$ 0.83
& 59.52 $\pm$ 0.19
& 66.50 $\pm$ 0.36 \\

& Base+CoT
& \cellcolor{pos10}67.10 $\pm$ 1.26
& \cellcolor{pos2}73.85 $\pm$ 0.54
& \cellcolor{pos8}71.08 $\pm$ 0.43
& \cellcolor{pos6}64.93 $\pm$ 0.54
& \cellcolor{pos6}71.27 $\pm$ 0.41 \\

& Base+FS
& \cellcolor{pos4}53.72 $\pm$ 0.92
& \cellcolor{pos1}73.73 $\pm$ 0.39
& \cellcolor{pos6}67.93 $\pm$ 0.30
& \cellcolor{pos7}66.97 $\pm$ 0.43
& \cellcolor{pos6}72.69 $\pm$ 0.14 \\

& Base+CoT+FS
& \cellcolor{pos9}62.63 $\pm$ 2.21
& \cellcolor{pos2}73.46 $\pm$ 0.43
& \cellcolor{pos8}71.41 $\pm$ 0.62
& \cellcolor{pos7}67.29 $\pm$ 0.36
& \cellcolor{pos7}72.51 $\pm$ 0.13 \\

& Base+CoS
& \cellcolor{neg1}49.68 $\pm$ 0.26
& \cellcolor{neg9}60.43 $\pm$ 1.05
& \cellcolor{pos1}61.47 $\pm$ 0.25
& \cellcolor{pos1}59.54 $\pm$ 1.24
& \cellcolor{neg7}57.96 $\pm$ 0.75 \\

& OPRO
& \cellcolor{pos9}62.82 $\pm$ 0.68
& \cellcolor{neg4}69.43 $\pm$ 0.89
& \cellcolor{neg1}60.59 $\pm$ 0.80
& \cellcolor{neg3}57.37 $\pm$ 0.86
& \cellcolor{neg1}66.17 $\pm$ 0.94 \\

& CAPO
& \cellcolor{pos9}62.89 $\pm$ 0.30
& \cellcolor{pos1}73.83 $\pm$ 0.36
& \cellcolor{pos3}64.10 $\pm$ 1.01
& \cellcolor{pos3}61.85 $\pm$ 1.13
& \cellcolor{neg1}66.49 $\pm$ 1.61 \\

& EvoPrompt(DE)
& \cellcolor{pos9}62.60 $\pm$ 2.27
& \cellcolor{neg2}72.08 $\pm$ 0.41
& \cellcolor{pos4}64.50 $\pm$ 1.68
& \cellcolor{pos6}65.05 $\pm$ 2.54
& \cellcolor{neg1}66.38 $\pm$ 1.62 \\

& EvoPrompt(GA)
& \cellcolor{pos9}64.69 $\pm$ 2.79
& \cellcolor{neg1}72.33 $\pm$ 1.48
& \cellcolor{pos4}65.00 $\pm$ 0.73
& \cellcolor{pos6}66.43 $\pm$ 1.30
& \cellcolor{pos1}67.31 $\pm$ 1.75 \\

& I (ours)
& \cellcolor{pos8}60.82 $\pm$ 1.06
& \cellcolor{neg1}72.50 $\pm$ 0.32
& \cellcolor{pos6}67.36 $\pm$ 1.23
& \cellcolor{pos6}64.56 $\pm$ 1.64
& \cellcolor{neg2}64.80 $\pm$ 1.30 \\

& I+R (ours)
& \cellcolor{pos10}\textbf{72.26 $\pm$ 1.77}
& \cellcolor{pos2}\textbf{74.79 $\pm$ 1.56}
& \cellcolor{pos9}73.99 $\pm$ 2.79
& \cellcolor{pos9}\textbf{73.01 $\pm$ 2.28}
& \cellcolor{pos6}71.78 $\pm$ 2.62 \\

& I+R+E (ours)
& \cellcolor{pos10}69.97 $\pm$ 2.43
& \cellcolor{neg3}71.14 $\pm$ 1.89
& \cellcolor{pos10}\textbf{78.03 $\pm$ 2.24}
& \cellcolor{pos8}69.55 $\pm$ 1.96
& \cellcolor{pos7}\textbf{73.58 $\pm$ 3.29} \\

\midrule

\multirow{12}{*}{VaccineStance}
& Base (zero-shot)
& 49.47 $\pm$ 0.31
& 58.84 $\pm$ 1.37
& 55.58 $\pm$ 0.69
& 37.78 $\pm$ 0.24
& 60.23 $\pm$ 0.47 \\

& Base+CoT
& \cellcolor{pos6}54.93 $\pm$ 0.56
& \cellcolor{pos2}59.46 $\pm$ 0.35
& \cellcolor{pos6}59.98 $\pm$ 0.74
& \cellcolor{pos10}52.36 $\pm$ 0.25
& \cellcolor{pos2}60.80 $\pm$ 0.32 \\

& Base+FS
& \cellcolor{pos1}49.88 $\pm$ 1.00
& \cellcolor{pos4}62.37 $\pm$ 0.60
& \cellcolor{pos8}65.24 $\pm$ 0.24
& \cellcolor{pos10}54.26 $\pm$ 0.25
& \cellcolor{pos1}60.51 $\pm$ 0.13 \\

& Base+CoT+FS
& \cellcolor{pos9}62.94 $\pm$ 0.30
& \cellcolor{pos5}63.53 $\pm$ 0.94
& \cellcolor{pos7}63.95 $\pm$ 0.44
& \cellcolor{pos10}62.72 $\pm$ 0.37
& \cellcolor{pos3}\textbf{62.45 $\pm$ 0.20} \\

& Base+CoS
& \cellcolor{pos5}53.83 $\pm$ 0.54
& \cellcolor{pos3}61.14 $\pm$ 0.65
& \cellcolor{pos7}63.18 $\pm$ 0.96
& \cellcolor{pos10}58.89 $\pm$ 0.87
& \cellcolor{neg6}54.48 $\pm$ 0.43 \\

& OPRO
& \cellcolor{pos5}53.48 $\pm$ 6.45
& \cellcolor{pos4}62.59 $\pm$ 0.60
& \cellcolor{pos4}58.80 $\pm$ 3.15
& \cellcolor{pos8}48.53 $\pm$ 1.29
& \cellcolor{neg4}56.90 $\pm$ 3.93 \\

& CAPO
& \cellcolor{pos7}57.37 $\pm$ 0.41
& \cellcolor{pos3}61.76 $\pm$ 0.20
& \cellcolor{pos4}59.15 $\pm$ 0.99
& \cellcolor{pos10}54.18 $\pm$ 4.76
& \cellcolor{neg2}58.41 $\pm$ 0.41 \\

& EvoPrompt(DE)
& \cellcolor{pos7}58.04 $\pm$ 1.07
& \cellcolor{pos2}60.74 $\pm$ 0.26
& \cellcolor{pos1}55.73 $\pm$ 2.17
& \cellcolor{pos7}46.26 $\pm$ 3.24
& \cellcolor{neg3}58.04 $\pm$ 0.37 \\

& EvoPrompt(GA)
& \cellcolor{pos6}56.45 $\pm$ 8.25
& \cellcolor{pos4}62.67 $\pm$ 1.80
& \cellcolor{pos1}55.66 $\pm$ 5.66
& \cellcolor{pos10}55.00 $\pm$ 1.03
& \cellcolor{neg4}56.94 $\pm$ 0.74 \\

& I (ours)
& \cellcolor{pos6}54.83 $\pm$ 5.98
& \cellcolor{pos4}62.74 $\pm$ 0.52
& \cellcolor{pos4}58.90 $\pm$ 2.88
& \cellcolor{pos3}40.02 $\pm$ 3.77
& \cellcolor{pos1}60.39 $\pm$ 1.65 \\

& I+R (ours)
& \cellcolor{pos8}60.21 $\pm$ 0.06
& \cellcolor{pos6}65.35 $\pm$ 1.61
& \cellcolor{pos9}\textbf{68.96 $\pm$ 2.00}
& \cellcolor{pos10}\textbf{66.00 $\pm$ 1.72}
& \cellcolor{neg2}59.06 $\pm$ 1.74 \\

& I+R+E (ours)
& \cellcolor{pos9}\textbf{63.87 $\pm$ 3.57}
& \cellcolor{pos7}\textbf{66.99 $\pm$ 1.20}
& \cellcolor{pos9}67.85 $\pm$ 3.59
& \cellcolor{pos10}55.12 $\pm$ 1.83
& \cellcolor{neg1}59.76 $\pm$ 1.27 \\

\bottomrule
\end{tabular}

\caption{
Average Macro-F1 across targets for each dataset, model, and prompting
strategy. I = instruction-only; I+R = instruction with reasoning;
I+R+E = instruction with per-example reasoning. Base = human-written
prompt; CoT = chain-of-thought prompting; FS = few-shot prompting; CoS =
Chain-of-Stance prompting. Results are reported as mean $\pm$ standard
deviation across three independent runs. Shaded cells indicate differences
relative to the Base score within each dataset--model column: gray denotes
improvements, red denotes decreases, and darker shades indicate larger
absolute differences. The best score within each dataset--model column is
shown in \textbf{bold}.
}
\label{tab:aggregate_results}

\end{table*}
\section{Results and Discussion}

We evaluate MoPrO-SD from both performance and diagnostic perspectives.
We first examine optimization dynamics and aggregate Macro-F1 performance
across datasets and models, followed by comparisons with structured prompt
optimization methods, external-reasoner effects, computational cost,
modular versus full-prompt optimization, and module-level ablations.

\subsection{Optimization Dynamics}

\begin{figure*}[!t]
    \centering
    \includegraphics[width=\textwidth]{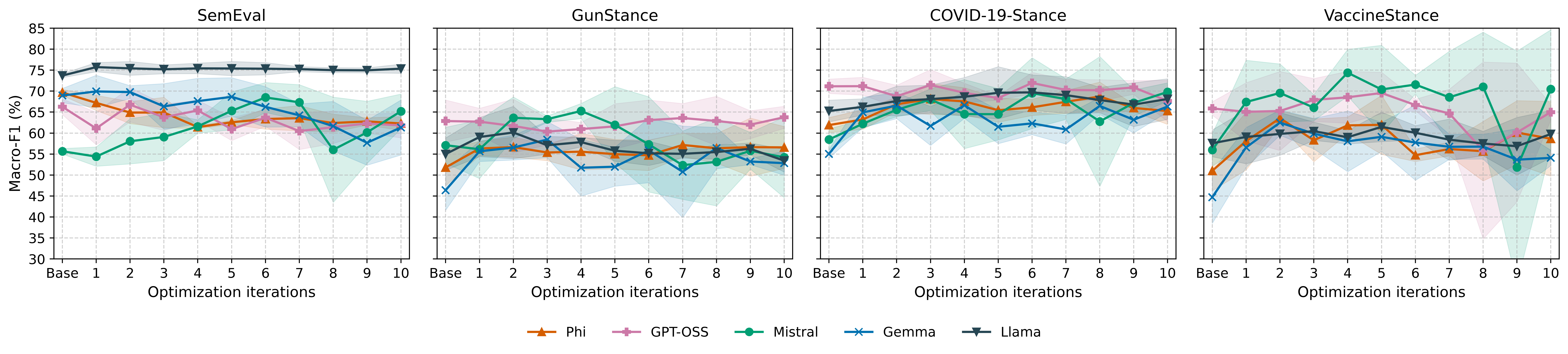}
    \caption{Validation Macro-F1 trajectories across 10 optimization
    iterations using I+R+E. Curves show mean validation Macro-F1 across
    three independent runs, with shaded regions indicating the corresponding
    standard deviation.}
    \label{fig:ire_trajectory}
\end{figure*}

Figure~\ref{fig:ire_trajectory} shows validation Macro-F1 trajectories
across 10 optimization iterations using I+R+E. Most models improve over
the Base prompt during the early iterations, while later iterations show
smaller gains and greater fluctuations. Some trajectories exhibit sharp
drops followed by recovery, particularly on VaccineStance, indicating
variation across optimization iterations. The shaded regions also show
run-to-run variability in validation performance. Overall, the
trajectories show that useful refinements are often identified early and
that continued optimization does not guarantee monotonic improvement.
The trajectories for I and I+R are provided
in Appendix~\ref{app:trajectory_plots}.

\subsection{Main Performance Results}

Table~\ref{tab:aggregate_results} summarizes Macro-F1 across datasets,
backbone models, and prompting strategies, with per-target results
provided in Appendix~\ref{app:detailed_results}. I+R+E improves over the
human-written Base prompt in \textbf{18 of 20} dataset--model combinations.
The improvements are nevertheless model- and dataset-dependent: I+R
outperforms I+R+E in several settings, indicating that the preferred
reasoning-based refinement configuration depends on the model and dataset.

The largest I+R+E gains over Base are observed for Phi on GunStance ($39.55 \rightarrow 59.74$, $+20.19$ Macro-F1), Phi on COVID-19-Stance ($50.36 \rightarrow 69.97$, $+19.61$), and Gemma on VaccineStance ($37.78 \rightarrow 55.12$, $+17.34$). Across all methods, CAPO with Llama obtains the highest SemEval score (74.87), while I+R+E achieves 74.78. I+R+E with Mistral also performs strongly on GunStance (65.01) and achieves the highest score on COVID-19-Stance (78.03), while I+R with Mistral obtains the highest VaccineStance score (68.96).

Static prompting also yields substantial improvements in some settings.
For example, Base+FS increases Mistral from 53.90 to 64.69 on SemEval
and from 55.58 to 65.24 on VaccineStance. MoPrO-SD frequently improves
further, suggesting that iterative module refinement can provide gains
beyond static prompt augmentation. However, the variation across models,
datasets, and refinement strategies indicates that no single optimization
strategy dominates in every setting. Additional zero-shot and few-shot
comparisons are provided in Appendix~\ref{app:fewshot_analysis}.

\paragraph{Comparison with Structured Prompt Optimization.}
We further compare MoPrO-SD with UniPrompt and SCULPT, two related
structured prompt optimization methods. As shown in
Table~\ref{tab:structured_prompt_comparison}, I+R+E achieves 71.12,
69.68, and 74.78 Macro-F1 with Mistral, Gemma, and Llama, respectively.
Its performance is similar to both UniPrompt settings on Mistral and
Gemma, while achieving the highest numerical score on Llama (74.78).
Overall, MoPrO-SD performs comparably to recent structured prompt
optimization approaches; computational cost is examined separately in
Section~\ref{sec:cost}.

\begin{table}[!t]
\centering

\begingroup
\setlength{\tabcolsep}{3.0pt}
\renewcommand{\arraystretch}{1.02}

\resizebox{\columnwidth}{!}{%
\begin{tabular}{lccc}
\toprule
\textbf{Method} &
\textbf{Mistral} &
\textbf{Gemma} &
\textbf{Llama} \\
\midrule

UniPrompt (Original Init.)
& $71.06 \pm 0.68$
& $69.82 \pm 0.62$
& $73.99 \pm 0.11$ \\

UniPrompt (Shared Init.)
& $\mathbf{71.42 \pm 1.94}$
& $\mathbf{69.90 \pm 2.49}$
& $73.33 \pm 0.73$ \\

SCULPT
& $53.93 \pm 0.40$
& $65.41 \pm 0.01$
& $74.19 \pm 0.48$ \\

\midrule

I+R+E (Ours)
& $71.12 \pm 0.85$
& $69.68 \pm 1.14$
& $\mathbf{74.78 \pm 0.38}$ \\

\bottomrule
\end{tabular}%
}

\endgroup

\caption{Comparison with structured prompt optimization methods on
SemEval-2016. Results are Macro-F1 (\%), reported as mean $\pm$
standard deviation across runs. Exploratory two-sided Welch tests are
unadjusted for multiple comparisons; Mistral--SCULPT gives $p<0.001$
and Gemma--SCULPT $p=0.023$, with all others $p>0.05$.}
\label{tab:structured_prompt_comparison}

\end{table}

\paragraph{External Reasoner Comparison.}
We evaluate GPT-5~\citep{openai2025gpt5} as an external reasoner
for the three label-specific I+R+E modules, while retaining
the target backbone for classification and other module updates.
This tests whether stronger reasoning during optimization
improves prompts without replacing the target classifier.
Table~\ref{tab:gpt5-optimizer-comparison} shows gains in
\textbf{17 of 20} dataset--model combinations, with
substantial variation across settings.

The largest gains occur for Gemma on VaccineStance (+11.64),
Llama on GunStance (+9.42), and Llama on COVID-19-Stance
(+9.23). However, Mistral on SemEval declines by 5.82 points,
showing that stronger reasoning does not always improve
optimization. Overall, reasoning quality can substantially
affect optimization outcomes, but its benefits remain
model- and dataset-dependent.
\begin{table}[!t]
\centering

\begingroup
\setlength{\tabcolsep}{3.0pt}
\renewcommand{\arraystretch}{1.02}

\resizebox{\columnwidth}{!}{%
\begin{tabular}{lrrrrr}
\toprule
& \multicolumn{5}{c}{\textbf{GPT-5 vs.\ Same LLM ($\Delta$ Macro-F1)}} \\
\cmidrule(lr){2-6}

\textbf{Dataset}
& \textbf{Phi}
& \textbf{GPT-OSS}
& \textbf{Mistral}
& \textbf{Gemma}
& \textbf{Llama} \\
\midrule

SemEval
& +0.48
& +0.25
& $-$5.82
& +2.23
& +1.68 \\

GunStance
& +1.87
& $-$1.56
& $-$0.87
& +2.78
& +9.42 \\

COVID-19
& +0.10
& +0.07
& +1.04
& +8.44
& +9.23 \\

VaccineStance
& +2.27
& +1.02
& +3.66
& \textbf{+11.64}
& +8.55 \\

\bottomrule
\end{tabular}%
}

\endgroup

\caption{Effect of using GPT-5 as the external reasoner for the
label-specific modules in I+R+E. Values report $\Delta$ Macro-F1 =
GPT-5 $-$ Same LLM. Full results are in
Appendix~\ref{app:gpt5_reasoner}.}
\label{tab:gpt5-optimizer-comparison}

\end{table}

\subsection{Token Cost Analysis}
\label{sec:cost}

\begin{table}[t]
\centering
\scriptsize
\setlength{\tabcolsep}{2.2pt}
\renewcommand{\arraystretch}{1.06}

\resizebox{\columnwidth}{!}{%
\begin{tabular}{lrrrrr}
\toprule
\textbf{Method}
& \textbf{Calls}
& \textbf{In.}
& \textbf{Out.}
& \textbf{Total}
& \textbf{Rel.} \\
\midrule

Base
& 1,249 & 768.9K & 4.4K & 773.4K & 1.00$\times$ \\

Base+FS
& 1,249 & 1.14M & 4.3K & 1.14M & 1.48$\times$ \\

Base+CoT
& 2,521 & 1.85M & 423.4K & 2.27M & 2.94$\times$ \\

Base+CoT+FS
& 2,520 & 2.60M & 420.8K & 3.02M & 3.91$\times$ \\

SCULPT
& 9,680 & 6.08M & 306.5K & 6.38M & 8.26$\times$ \\

I+R+E (FP-5)
& 7,851 & 6.25M & 133.8K & 6.39M & 8.26$\times$ \\

I
& 7,741 & 6.14M & 376.9K & 6.52M & 8.43$\times$ \\

Base+CoS
& 7,494 & 5.96M & 1.24M & 7.19M & 9.30$\times$ \\

I+R+E (FP-BM)
& 9,563 & 10.98M & 142.7K & 11.12M & 14.38$\times$ \\

I+R
& 9,878 & 13.67M & 332.9K & 14.00M & 18.10$\times$ \\

OPRO
& 36,035 & 17.03M & 181.4K & 17.21M & 22.25$\times$ \\

I+R+E (Mod.)
& 12,568 & 18.34M & 1.16M & 19.50M & 25.21$\times$ \\

UniPrompt-O
& 36,944 & 15.84M & 8.47M & 24.31M & 31.43$\times$ \\

UniPrompt-S
& 30,211 & 21.49M & 3.30M & 24.79M & 32.05$\times$ \\

CAPO
& 48,224 & 29.80M & 363.2K & 30.17M & 39.01$\times$ \\

EvoPrompt-GA
& 53,437 & 28.79M & 2.83M & 31.61M & 40.88$\times$ \\

EvoPrompt-DE
& 40,972 & 39.72M & 7.50M & 47.22M & 61.05$\times$ \\

\bottomrule
\end{tabular}%
}

\caption{
LLM cost on SemEval-2016 using Mistral for one run across five targets.
Costs include all algorithmic LLM calls under each evaluated protocol,
including one final test pass. Relative cost is normalized by Base
(773,391 tokens). FP-5 and FP-BM denote five-iteration and best-only-memory
full-prompt I+R+E; Mod.\ denotes modular I+R+E.
}
\label{tab:cost_analysis}
\end{table}

As shown in Table~\ref{tab:cost_analysis}, I, I+R,
and modular I+R+E consume 8.43$\times$, 18.10$\times$,
and 25.21$\times$ the Base token budget, respectively.
Modular I+R+E uses fewer tokens than both UniPrompt
variants, CAPO, and both EvoPrompt variants, but more
than OPRO and the evaluated full-prompt variants. These costs reflect prompt development and evaluation
under the specified protocol; once optimized, the selected
prompt requires no additional optimization calls at inference.

\subsection{Ablation Studies}
\label{sec:full_prompt_ablation}

\paragraph{Modular vs. Full-Prompt Optimization.}
Table~\ref{tab:ire_full_prompt_ablation} compares modular I+R+E with full-prompt optimization on SemEval-2016. Modular optimization achieves higher mean Macro-F1 for four of five backbones, with the largest gain for Mistral (+3.49), while Llama remains nearly unchanged (+0.05) and Gemma favors full-prompt optimization ($-0.85$).

\begin{table*}[!t]
\centering
\small
\setlength{\tabcolsep}{5pt}
\renewcommand{\arraystretch}{1.08}
\fontsize{6.5}{7.5}\selectfont

\begin{tabular}{lccccc}
\toprule
\textbf{Method}
& \textbf{Phi}
& \textbf{GPT-OSS}
& \textbf{Mistral}
& \textbf{Gemma}
& \textbf{Llama} \\
\midrule

Base
& 55.37 $\pm$ 0.12
& 60.02 $\pm$ 0.44
& 53.90 $\pm$ 0.28
& 65.82 $\pm$ 0.20
& 73.52 $\pm$ 0.14 \\

I+R+E (Full Prompt, 5 iter.)
& \cellcolor{pos8}65.69 $\pm$ 1.88
& \cellcolor{pos6}65.46 $\pm$ 0.62
& \cellcolor{pos9}67.02 $\pm$ 1.08
& \cellcolor{pos6}\textbf{70.82 $\pm$ 0.39}
& \cellcolor{pos1}74.03 $\pm$ 0.83 \\

I+R+E (Full Prompt, Best-Only Memory, 10 iter.)
& \cellcolor{pos8}66.67 $\pm$ 0.76
& \cellcolor{pos7}65.10 $\pm$ 0.94
& \cellcolor{pos9}67.63 $\pm$ 0.61
& \cellcolor{pos5}70.53 $\pm$ 0.58
& \cellcolor{pos2}74.73 $\pm$ 0.77 \\

I+R+E (Modular, 10 iter.)
& \cellcolor{pos9}\textbf{68.63 $\pm$ 0.53}
& \cellcolor{pos6}\textbf{66.10 $\pm$ 1.14}
& \cellcolor{pos10}\textbf{71.12 $\pm$ 0.85}
& \cellcolor{pos4}69.68 $\pm$ 1.14
& \cellcolor{pos2}\textbf{74.78 $\pm$ 0.38} \\

\bottomrule
\end{tabular}

\caption{
Comparison of full-prompt and modular I+R+E optimization on SemEval-2016
(Macro-F1, mean $\pm$ standard deviation over three runs).
Standard full-prompt optimization uses five iterations due to context
growth; the best-only-memory and modular variants use ten.
Bold indicates the best optimized result for each backbone.
}
\label{tab:ire_full_prompt_ablation}
\end{table*}

\paragraph{Module Ablation.}
\begin{table*}[!t]
\centering
\begingroup
\fontsize{7.2}{8.2}\selectfont
\setlength{\tabcolsep}{4.5pt}
\renewcommand{\arraystretch}{1.12}

\begin{adjustbox}{max width=\textwidth}
\begin{tabular}{lcc cc cc cc cc}
\toprule
\multirow{2}{*}{\textbf{Frozen Module}}
& \multicolumn{2}{c}{\textbf{Phi}}
& \multicolumn{2}{c}{\textbf{GPT-OSS}}
& \multicolumn{2}{c}{\textbf{Mistral}}
& \multicolumn{2}{c}{\textbf{Gemma}}
& \multicolumn{2}{c}{\textbf{Llama}} \\

\cmidrule(lr){2-3}
\cmidrule(lr){4-5}
\cmidrule(lr){6-7}
\cmidrule(lr){8-9}
\cmidrule(lr){10-11}

& \textbf{Macro-F1} & $\boldsymbol{\Delta}$\textbf{F1}
& \textbf{Macro-F1} & $\boldsymbol{\Delta}$\textbf{F1}
& \textbf{Macro-F1} & $\boldsymbol{\Delta}$\textbf{F1}
& \textbf{Macro-F1} & $\boldsymbol{\Delta}$\textbf{F1}
& \textbf{Macro-F1} & $\boldsymbol{\Delta}$\textbf{F1} \\
\midrule

None (All Modules Optimized)
& 68.63 $\pm$ 0.53 & --
& 66.10 $\pm$ 1.14 & --
& 71.12 $\pm$ 0.85 & --
& 69.68 $\pm$ 1.14 & --
& 74.78 $\pm$ 0.38 & -- \\

Favor definition
& \cellcolor{neg8}59.29 $\pm$ 2.32
& \cellcolor{neg8}$-9.34$
& \cellcolor{neg6}59.93 $\pm$ 2.77
& \cellcolor{neg6}$-6.17$
& \cellcolor{neg7}63.39 $\pm$ 1.43
& \cellcolor{neg7}$-7.73$
& \cellcolor{neg5}65.18 $\pm$ 1.07
& \cellcolor{neg5}$-4.50$
& \cellcolor{neg1}73.88 $\pm$ 0.90
& \cellcolor{neg1}$-0.90$ \\

Neutral definition
& \cellcolor{neg6}61.74 $\pm$ 1.66
& \cellcolor{neg6}$-6.89$
& \cellcolor{neg1}65.16 $\pm$ 1.24
& \cellcolor{neg1}$-0.94$
& \cellcolor{neg4}67.52 $\pm$ 0.41
& \cellcolor{neg4}$-3.60$
& \cellcolor{neg2}68.32 $\pm$ 1.62
& \cellcolor{neg2}$-1.36$
& \cellcolor{neg1}74.15 $\pm$ 0.39
& \cellcolor{neg1}$-0.63$ \\

Task definition
& \cellcolor{neg7}61.49 $\pm$ 1.91
& \cellcolor{neg7}$-7.14$
& \cellcolor{neg4}62.45 $\pm$ 1.35
& \cellcolor{neg4}$-3.65$
& \cellcolor{neg6}64.96 $\pm$ 0.73
& \cellcolor{neg6}$-6.16$
& \cellcolor{neg2}67.82 $\pm$ 0.64
& \cellcolor{neg2}$-1.86$
& \cellcolor{neg2}73.23 $\pm$ 1.21
& \cellcolor{neg2}$-1.55$ \\

Guidelines
& \cellcolor{neg4}65.09 $\pm$ 2.39
& \cellcolor{neg4}$-3.54$
& \cellcolor{neg4}62.97 $\pm$ 0.34
& \cellcolor{neg4}$-3.13$
& \cellcolor{neg6}65.46 $\pm$ 2.76
& \cellcolor{neg6}$-5.66$
& \cellcolor{neg1}69.17 $\pm$ 1.03
& \cellcolor{neg1}$-0.51$
& \cellcolor{neg2}73.00 $\pm$ 1.04
& \cellcolor{neg2}$-1.78$ \\

Against definition
& \cellcolor{neg7}61.02 $\pm$ 2.41
& \cellcolor{neg7}$-7.61$
& \cellcolor{neg6}59.51 $\pm$ 1.46
& \cellcolor{neg6}$-6.59$
& \cellcolor{neg8}61.46 $\pm$ 2.53
& \cellcolor{neg8}$-9.66$
& \cellcolor{neg2}68.02 $\pm$ 1.28
& \cellcolor{neg2}$-1.66$
& \cellcolor{neg2}73.22 $\pm$ 0.65
& \cellcolor{neg2}$-1.56$ \\

Output format
& \cellcolor{neg7}60.40 $\pm$ 4.09
& \cellcolor{neg7}$-8.23$
& \cellcolor{neg3}63.18 $\pm$ 1.09
& \cellcolor{neg3}$-2.92$
& \cellcolor{neg5}66.64 $\pm$ 2.28
& \cellcolor{neg5}$-4.48$
& \cellcolor{pos1}70.10 $\pm$ 0.75
& \cellcolor{pos1}$+0.42$
& \cellcolor{pos1}75.11 $\pm$ 0.51
& \cellcolor{pos1}$+0.33$ \\

\bottomrule
\end{tabular}
\end{adjustbox}

\caption{
Frozen-module ablation on SemEval-2016 under I+R+E optimization
(Macro-F1, mean $\pm$ standard deviation over three runs).
Each row freezes one module while optimizing the remainder.
$\Delta$F1 is relative to optimizing all modules.
}
\label{tab:module-ablation}

\endgroup
\end{table*}

Table~\ref{tab:module-ablation} presents the frozen-module ablation
results on SemEval-2016 under I+R+E optimization, where each experiment
freezes one module at its base-prompt version while optimizing the
remaining modules.

Freezing the Favor definition produces the largest degradation for Phi
($-9.34$) and Gemma ($-4.50$), whereas freezing the Against definition
has the largest effect for GPT-OSS ($-6.59$) and Mistral ($-9.66$).
Llama is comparatively less sensitive, with its largest degradation
($-1.78$) occurring when Guidelines is frozen. Notably, the Output
format is the only module for which freezing does not consistently
reduce performance, yielding small improvements for Gemma ($+0.42$)
and Llama ($+0.33$). Overall, these results indicate that the
contribution of individual prompt modules varies substantially across
backbone models. Qualitative examples are provided in
Appendix~\ref{app:frozen_examples}.

\paragraph{Class-Conditioned Error Routing.}
We compare I+R+E with a pooled-error variant using the same
initial prompt, ten iterations, and training batches of up to
45 examples. Validation subsets are fixed within each run and
shared across variants with the same run ID. Class-conditioned
routing uses all errors of the corresponding gold label,
whereas pooled routing samples an equal number across labels.
Modules with no class-specific errors remain unchanged.
Table~\ref{tab:routing_ablation} shows higher mean Macro-F1
for four of five backbones: Phi (+3.40), GPT-OSS (+1.05),
Mistral (+4.99), and Gemma (+2.21), while Llama decreases
by 0.15 points.

\begin{table}[!h]
\centering
\begingroup
\setlength{\tabcolsep}{5.5pt}
\renewcommand{\arraystretch}{1.05}

\resizebox{\columnwidth}{!}{%
\begin{tabular}{lccccc}
\toprule
\textbf{Routing} &
\textbf{Phi} &
\textbf{GPT} &
\textbf{Mistral} &
\textbf{Gemma} &
\textbf{Llama} \\
\midrule

Pooled
& $65.23 \pm 3.87$
& $65.05 \pm 1.01$
& $66.13 \pm 1.94$
& $67.47 \pm 1.48$
& $\mathbf{74.93 \pm 0.89}$ \\

Class-conditioned
& $\mathbf{68.63 \pm 0.53}$
& $\mathbf{66.10 \pm 1.14}$
& $\mathbf{71.12 \pm 0.85}$
& $\mathbf{69.68 \pm 1.14}$
& $74.78 \pm 0.38$ \\

\bottomrule
\end{tabular}%
}

\caption{Class-conditioned versus pooled error routing on SemEval-2016.
Values are mean Macro-F1 $\pm$ standard deviation over three runs,
averaged across five targets.}
\label{tab:routing_ablation}

\endgroup
\end{table}
\subsection{Error Analysis}
\label{sec:error_analysis}

We qualitatively examine cases where optimization corrects Base-prompt
errors. The examples illustrate improvements on challenging phenomena
such as implicit stance, sarcasm, and ambiguous expressions. Additional
cases are provided in Appendix~\ref{sec:qualitative_case_studies}.

\begin{casebox}{Case Study: Sarcastic Stance}
\small
\textbf{Dataset:} SemEval-2016 \\
\textbf{Target:} Atheism

\vspace{3pt}
\textbf{Tweet:}
\emph{``Irony: When a \#creationist tweets about \#science being wrong on
a device that couldn't have existed without it. \#SemST''}

\vspace{3pt}
\textbf{Gold:} \labeltag{correctbg}{\it favor}
\quad
\textbf{Base:} \labeltag{errorbg}{\it against}
\quad
\textbf{I:} \labeltag{correctbg}{\it favor}
\quad
\textbf{I+R:} \labeltag{errorbg}{\it against}
\quad
\textbf{I+R+E:} \labeltag{correctbg}{\it favor}

\vspace{3pt}
\textbf{Analysis:}
The tweet uses sarcasm to criticize a creationist position, requiring
the stance toward the target to be inferred beyond surface polarity.
The Base and I+R prompts predict \emph{against}, whereas I and I+R+E
recover the gold \emph{favor} label.
\end{casebox}

\section{Conclusion}

We introduced MoPrO-SD, a modular prompt optimization framework for
stance detection that refines prompt components without model fine-tuning.
Across four datasets and multiple LLM backbones, MoPrO-SD improves over
the Base prompt in many settings, although performance varies by model
and refinement strategy.

Class-conditioned error feedback achieves higher mean Macro-F1 than
pooled feedback for four of five backbones on SemEval-2016, supporting
the value of label-specific refinement. Modular optimization also
outperforms full-prompt optimization for several models, with varying
module contributions. Despite higher development-time costs, MoPrO-SD
requires no additional optimization at inference.
\clearpage

\section*{Limitations}

MoPrO-SD has several limitations. First, although it improves performance across multiple stance detection datasets, its generalizability to other NLP tasks remains unexplored. Second, optimized prompts are model- and domain-dependent, so transferring them across backbone models or datasets may require re-optimization. Third, the current framework uses a fixed modular structure tailored to stance detection, with explicit \textit{favor}, \textit{against}, and \textit{neutral} definition modules; extending this design to tasks with larger label spaces or more complex reasoning requirements remains future work. Fourth, MoPrO-SD incurs additional optimization-time cost due to repeated validation and optimizer calls, especially for reasoning-based strategies. Finally, experiments involving API-based LLMs may be affected by model updates over time, which can impact reproducibility.

\bibliography{references}

\clearpage
\appendix

\setcounter{page}{1}
\setcounter{section}{0}
\setcounter{table}{0}
\setcounter{figure}{0}
\newcounter{offset}
  \setcounter{offset}{\value{figure}}
  \renewcommand{\thefigure}{S\the\numexpr\value{figure}-\value{offset}\relax}
  \setcounter{offset}{\value{table}}
  \renewcommand{\thetable}{T\the\numexpr\value{table}-\value{offset}\relax}


\onecolumn
\section{Modular Prompt Design and Meta-Prompts}
\label{app:prompt design}

\subsection{Complete Prompt Template}
\label{sec:complete_prompt}

\begin{tcolorbox}[
enhanced,
breakable,
colback=white,
colframe=black!70,
boxrule=0.8pt,
arc=2pt,
left=6pt,
right=6pt,
top=5pt,
bottom=4pt,
title=\textbf{Complete Base Prompt for Stance Detection},
fonttitle=\bfseries\small,
colbacktitle=gray!45,
coltitle=black,
boxed title style={
boxrule=0pt,
arc=1.5pt,
left=5pt,
right=5pt,
top=3pt,
bottom=3pt
}
]

\footnotesize


\vspace{1pt}
{\large\textbf{1. Task Definition}}

\vspace{3pt}
\noindent
Stance detection is a task of determining the attitude or position expressed by an author in a text regarding a particular topic or claim. Your task is to classify the stance of a tweet toward the target ``\texttt{\{target\}}'' as either Favor, Against, or Neutral.

\vspace{8pt}
\hrule
\vspace{8pt}


{\large\textbf{2. Guidelines}}

\vspace{4pt}

\begin{enumerate}
    \item \textbf{Stance Indicators:}
    Look for keywords, phrases, emoticons, and tone that indicate a favor, against, or neutral attitude toward \texttt{\{target\}}.

    \vspace{3pt}

    \item \textbf{Implicit vs. Explicit Stance:}
    Consider both implicit and explicit expressions of stance. If a tweet implies a stance without directly stating it, use the context and language to determine the intended stance.

    \vspace{3pt}

    \item \textbf{Sarcasm and Irony:}
    Be aware of sarcasm and irony, which can be used to express an opposing stance. Look for inconsistencies between the literal meaning and the tone or language used.
\end{enumerate}

\vspace{6pt}
\hrule
\vspace{8pt}


{\large\textbf{3. Favor Definition}}

\vspace{3pt}

Classify the tweet as favor if it:

\begin{enumerate}
    \item Shows an explicitly positive stance: openly supports \texttt{\{target\}}

    \vspace{2pt}

    \item Uses pro-\texttt{\{target\}} keywords: languages that frames \texttt{\{target\}} in positive light
\end{enumerate}

\vspace{6pt}
\hrule
\vspace{8pt}


{\large\textbf{4. Against Definition}}

\vspace{3pt}

Classify the tweet as against if it:

\begin{enumerate}
    \item Shows an explicitly negative stance: openly opposes \texttt{\{target\}}

    \vspace{2pt}

    \item Uses against \texttt{\{target\}} keywords: languages that frames the target \texttt{\{target\}} in negative light
\end{enumerate}

\vspace{6pt}
\hrule
\vspace{8pt}


{\large\textbf{5. Neutral Definition}}

\vspace{3pt}

Classify the tweet as neutral if it:

\begin{enumerate}
    \item Shows no clear stance: it neither supports nor opposes \texttt{\{target\}}

    \vspace{2pt}

    \item Uses neutral or tentative language: e.g., simply asking a question or stating opinion that signals indifference/uncertainty
\end{enumerate}

\vspace{6pt}
\hrule
\vspace{8pt}


{\large\textbf{6. Output Format}}

\vspace{3pt}

Strictly follow the guidelines above and classify the tweet if it is Favor, Against, or Neutral to the target ``\texttt{\{target\}}''. Reply with the classification label only, without any additional text or explanation.

\vspace{8pt}

\begin{tcolorbox}[
colback=black!2,
colframe=black!30,
boxrule=0.4pt,
arc=1.5pt,
left=4pt,
right=4pt,
top=4pt,
bottom=4pt
]

\texttt{Tweet: \{tweet\}}

\vspace{4pt}

\texttt{Stance: [Favor/Against/Neutral]}

\end{tcolorbox}

\end{tcolorbox}

\twocolumn
\subsection{Meta Prompts}
\label{sec:appendix_a_meta_prompts}


\definecolor{prompttitle}{RGB}{95,95,95}
\definecolor{promptborder}{RGB}{150,150,150}
\definecolor{promptbg}{RGB}{250,250,250}

\tcbset{
metaprompt/.style={
enhanced,
breakable,
colback=white,
colframe=gray!45,
coltitle=black,
colbacktitle=gray!45,
fonttitle=\bfseries\scriptsize,
boxrule=0.55pt,
arc=1.2pt,
left=3pt,
right=3pt,
top=2pt,
bottom=2pt,
toptitle=2.5pt,
bottomtitle=2.5pt,
},
promptinner/.style={
colback=promptbg,
colframe=promptborder!55,
boxrule=0pt,
arc=1pt,
left=2pt,
right=2pt,
top=1pt,
bottom=1pt
}
}


\begin{tcolorbox}[metaprompt,title=Instruction-Only Update]
\footnotesize

Your task is to write only the \texttt{\{module\_name\}} section of the
prompt for stance detection toward the target \texttt{\{target\}}.

\vspace{2pt}

Below are the previous prompts with their Macro-F1 scores. The Macro-F1
score ranges from 0 to 100, and higher scores indicate better-performing
prompts.

\vspace{2pt}

\begin{tcolorbox}[promptinner]
\texttt{\{prompt\_1\}}\\
\texttt{\{score\_1\}}\\
\texttt{...}\\
\texttt{\{prompt\_n\}}\\
\texttt{\{score\_n\}}
\end{tcolorbox}

\vspace{-2pt}

Carefully analyze the previous prompts and their Macro-F1 scores, and
write a new, improved prompt to replace the prompt below:

\vspace{-2pt}

\begin{tcolorbox}[promptinner]
\centering
\texttt{\{to\_rewrite\_prompt\}}
\end{tcolorbox}

\vspace{-2pt}

Wrap the new improved prompt between \texttt{START} and \texttt{END}
tags. Do not explain your changes or include any other text.

\end{tcolorbox}


\begin{tcolorbox}[metaprompt,title=Reasoning Step]
\footnotesize

Your task is to point out the problems with the current prompt based on
the wrong examples. The current prompt is:

\vspace{-2pt}

\begin{tcolorbox}[promptinner]
\texttt{\{current\_section\_prompt\}}
\end{tcolorbox}

\vspace{-2pt}

This prompt gets the following examples wrong. You should analyze the
differences between wrong predictions and ground truth answers, and
carefully consider why this prompt led to incorrect predictions.

\vspace{-2pt}

Below are the examples with Tweets, Wrong predictions, and Ground truth
answers.

\vspace{2pt}

\begin{tcolorbox}[promptinner]
\texttt{\{error\_demonstrations\}}
\end{tcolorbox}

\vspace{-1pt}

Give a reason why the prompt could have gotten these examples wrong, and
how we can improve the prompt? Wrap the reason with \texttt{START} and
\texttt{END} tags.

\end{tcolorbox}


\begin{tcolorbox}[metaprompt,title=Reasoning Update]
\footnotesize

Your task is to write an improved prompt based on the feedback. Below is
the current prompt with its Macro-F1 score. The Macro-F1 score ranges
from 0 to 100, and a higher score indicates better performance.

\vspace{2pt}

\begin{tcolorbox}[promptinner]
Prompt: \texttt{\{current\_prompt\}}\\
Macro-F1 Score: \texttt{\{current\_prompt\_score\}}
\end{tcolorbox}

\vspace{2pt}

Below are the problems with this prompt.

\vspace{2pt}

\begin{tcolorbox}[promptinner]
\texttt{\{problems\_reasoning\}}
\end{tcolorbox}

\vspace{2pt}

Carefully analyze the previous prompt and the problems with this prompt.
Write a concise, new, improved prompt. Wrap the new improved prompt
between \texttt{START} and \texttt{END} tags. Do NOT explain your changes
or include any other text between \texttt{START} and \texttt{END} tags.

\end{tcolorbox}


\begin{tcolorbox}[metaprompt,title=Per-Example Reasoning]
\footnotesize

Your task is to point out the problems with the current prompt based on
why this prompt leads to incorrect predictions. The current prompt is:

\vspace{2pt}

\begin{tcolorbox}[promptinner]
\texttt{\{current\_section\_prompt\}}
\end{tcolorbox}

\vspace{2pt}

You should analyze the differences between the wrong prediction and the
ground truth answer. Below is an example with a Tweet, a Wrong prediction,
and the Ground truth answer.

\vspace{2pt}

\begin{tcolorbox}[promptinner]
Tweet: \texttt{\{tweet\}}\\
Target: \texttt{\{target\}}\\
Ground Truth: \texttt{\{ground\_truth\}}\\
Wrong Prediction: \texttt{\{wrong\_prediction\}}
\end{tcolorbox}

\vspace{2pt}

Give a reason why the prompt could have gotten this example wrong, and
how we can improve the prompt? Wrap the reason with \texttt{START} and
\texttt{END} tags.

\end{tcolorbox}


\begin{tcolorbox}[metaprompt,title=Per-Example Reasoning Summarization]
\footnotesize

Your task is to summarize the problems with the current prompt and ways
to improve the prompt based on the problems defined below:

\vspace{-2pt}

\begin{tcolorbox}[promptinner]
\texttt{\{problem\_reasoning\}}
\end{tcolorbox}

\vspace{-2pt}

Write a short and concise summary between \texttt{START} and
\texttt{END} tags. Do NOT explain your changes or include any other text
between \texttt{START} and \texttt{END} tags.

\end{tcolorbox}


\begin{tcolorbox}[metaprompt,title=Post-Summary Module Update]
\footnotesize

Your task is to write an improved prompt based on the feedback. Below is
the current prompt:

\vspace{2pt}

\begin{tcolorbox}[promptinner]
\texttt{\{current\_section\_prompt\}}
\end{tcolorbox}

\vspace{2pt}

Below is an example with a Tweet, Wrong prediction, and Ground truth
answer.

\vspace{2pt}

\begin{tcolorbox}[promptinner]
Tweet: \texttt{\{tweet\}}\\
Target: \texttt{\{target\}}\\
Ground Truth: \texttt{\{ground\_truth\_answer\}}\\
Wrong Prediction: \texttt{\{wrong\_prediction\}}
\end{tcolorbox}

\vspace{2pt}

Below are the problems with this prompt.

\vspace{2pt}

\begin{tcolorbox}[promptinner]
\texttt{\{problems\_reasoning\}}
\end{tcolorbox}

\vspace{2pt}

Carefully analyze the previous prompt and the problems with this prompt.
Write a short, concise and improved prompt to replace the current prompt
without losing the important information. The current prompt is:

\vspace{2pt}

\begin{tcolorbox}[promptinner]
\texttt{\{current\_section\_prompt\}}
\end{tcolorbox}

\vspace{2pt}

Wrap the new improved prompt between \texttt{<START>} and
\texttt{<END>} tags. Do NOT explain your changes or include any other
text between \texttt{<START>} and \texttt{<END>} tags.

\end{tcolorbox}


\twocolumn

\section{Additional Experimental Details}
\label{appendix_additional_details}

\subsection{Dataset Statistics}

Table~\ref{tab:dataset_distribution} summarizes the target-wise
train, validation, and test distributions for the four datasets used
in our experiments.

\begin{table*}[t]
\centering
\fontsize{8.5}{9.5}\selectfont
\setlength{\tabcolsep}{4pt}
\renewcommand{\arraystretch}{0.88}

\begin{tabular}{llrrrr}
\toprule
\textbf{Dataset} & \textbf{Target} & \textbf{Train} & \textbf{Valid} & \textbf{Test} & \textbf{Total} \\
\midrule

\multirow{4}{*}{COVID-19-Stance}
& Anthony Fauci        & 1,464 & 200 & 200 & 1,864 \\
& Face Masks           & 1,307 & 200 & 200 & 1,707 \\
& School Closures      &   790 & 200 & 200 & 1,190 \\
& Stay at Home Orders  &   972 & 200 & 200 & 1,372 \\
\cmidrule(lr){2-6}
& \textbf{Total}       & \textbf{4,533} & \textbf{800} & \textbf{800} & \textbf{6,133} \\
\midrule

\multirow{2}{*}{GunStance}
& Guns Should Be Banned    & 900 & 900 & 900 & 2,700 \\
& Guns Should Be Regulated & 936 & 932 & 932 & 2,800 \\
\cmidrule(lr){2-6}
& \textbf{Total}           & \textbf{1,836} & \textbf{1,832} & \textbf{1,832} & \textbf{5,500} \\
\midrule

\multirow{5}{*}{SemEval-2016}
& Atheism                           & 389 & 124 & 220 & 733 \\
& Climate Change Is a Real Concern & 324 &  71 & 169 & 564 \\
& Feminist Movement                & 536 & 128 & 285 & 949 \\
& Hillary Clinton                  & 557 & 132 & 295 & 984 \\
& Legalization of Abortion         & 525 & 128 & 280 & 933 \\
\cmidrule(lr){2-6}
& \textbf{Total}                   & \textbf{2,331} & \textbf{583} & \textbf{1,249} & \textbf{4,163} \\
\midrule

\multirow{1}{*}{VaccineStance}
& Vaccines                & 500 & 500 & 1,000 & 2,000 \\
\cmidrule(lr){2-6}
& \textbf{Total}          & \textbf{500} & \textbf{500} & \textbf{1,000} & \textbf{2,000} \\

\bottomrule
\end{tabular}
\caption{Target-wise train, validation, and test distributions for the
stance detection benchmarks.}
\label{tab:dataset_distribution}
\end{table*}

\subsection{VaccineStance Construction and Annotation}
\label{app:vaccinestance}

We construct VaccineStance from the vaccine-discourse corpus
of Gyawali et al.~\cite{gyawali2025vaccine}, comprising public
English-language posts collected from X (formerly Twitter)
using the Full Archive Search API between January 2013 and
December 2022. Preprocessing retained original posts, limited
users to one post per day, removed reposts, URLs, emoticons,
non-ASCII characters, and user mentions, and filtered references
to the band ``The Vaccines'' using Sentence-Transformer similarity.
The resulting corpus contained approximately 18.7 million posts
from 5.8 million users.

From this corpus, we sampled 240,000 tweets (2,000 per month)
and used Llama-3.3~\cite{meta_llama3p3_70b_2024} to identify
posts relevant to vaccine stance detection. We then selected
2,000 tweets for human stance annotation. Each tweet was
independently labeled by three annotators as \textit{Favor},
\textit{Against}, or \textit{Neutral}, achieving Fleiss'
$\kappa=0.778$. The resulting benchmark contains 724 Favor
(36.2\%), 464 Against (23.2\%), and 812 Neutral (40.6\%)
instances, divided into 500 training, 500 validation, and
1,000 test instances. All tweet identifiers and texts are
unique, with no exact overlap across splits.

\subsection{Sample Annotated VaccineStance Examples}

Representative VaccineStance examples with Llama-3.3 stance predictions
and final human annotations are shown below. These stance predictions
are distinct from the relevance-filtering step used during candidate
selection.

\vspace{0.4em}

\begin{casebox}{Example 1: Neutral Agreement}
\small

\textbf{Post:}

\emph{``Parents: heard the new vaccine requirement? Meningitis vaccine
required for IL students entering 6th, 12th grade.''}

\vspace{3pt}

\textbf{Llama-3.3:}
\labeltag{correctbg}{\it neutral}
\quad
\textbf{Human:}
\labeltag{correctbg}{\it neutral}

\end{casebox}

\vspace{-0.55cm}

\begin{casebox}{Example 2: Against Agreement}
\small

\textbf{Post:}

\emph{``Then it's not recommended. The govt is making the public suffer
by gambling with this, I can't. Make the non-medical govt officials
take the vaccine first.''}

\vspace{3pt}

\textbf{Llama-3.3:}
\labeltag{correctbg}{\it against}
\quad
\textbf{Human:}
\labeltag{correctbg}{\it against}

\end{casebox}

\vspace{-0.35cm}

\begin{casebox}{Example 3: Annotation Disagreement}
\small

\textbf{Post:}

\emph{``One of the last holy grails of HIV research is the development
of a HIV vaccine.'' \#homenews \#news \#cryptonews}

\vspace{3pt}

\textbf{Llama-3.3:}
\labeltag{errorbg}{\it neutral}
\quad
\textbf{Human:}
\labeltag{correctbg}{\it in-favor}

\end{casebox}

\vspace{-0.35cm}

\begin{casebox}{Example 4: Implicit Support Disagreement}
\small

\textbf{Post:}

\emph{``Watching the age go up for what’s considered eligible for the
vaccine. KNOCK IT OFF.''}

\vspace{3pt}

\textbf{Llama-3.3:}
\labeltag{errorbg}{\it against}
\quad
\textbf{Human:}
\labeltag{correctbg}{\it in-favor}

\end{casebox}

\begin{table}[t]
\centering

\fontsize{8.5}{9.5}\selectfont
\setlength{\tabcolsep}{5pt}
\renewcommand{\arraystretch}{0.95}

\begin{tabular}{lrrrr}
\toprule
\textbf{Split} & \textbf{Favor} & \textbf{Against} &
\textbf{Neutral} & \textbf{Total} \\
\midrule
Train      & 159 & 103 & 238 & 500 \\
Validation & 169 & 102 & 229 & 500 \\
Test       & 396 & 259 & 345 & 1,000 \\
\midrule
\textbf{Total} &
\textbf{724} & \textbf{464} & \textbf{812} & \textbf{2,000} \\
\bottomrule
\end{tabular}

\caption{Class distribution of VaccineStance across the
train, validation, and test splits.}
\label{tab:vaccine_class_distribution}

\end{table}

\section{Effect of Optimization Budget}
\label{app:optimization_budget}

\begin{figure}[!ht]
    \centering
    \includegraphics[width=\columnwidth]{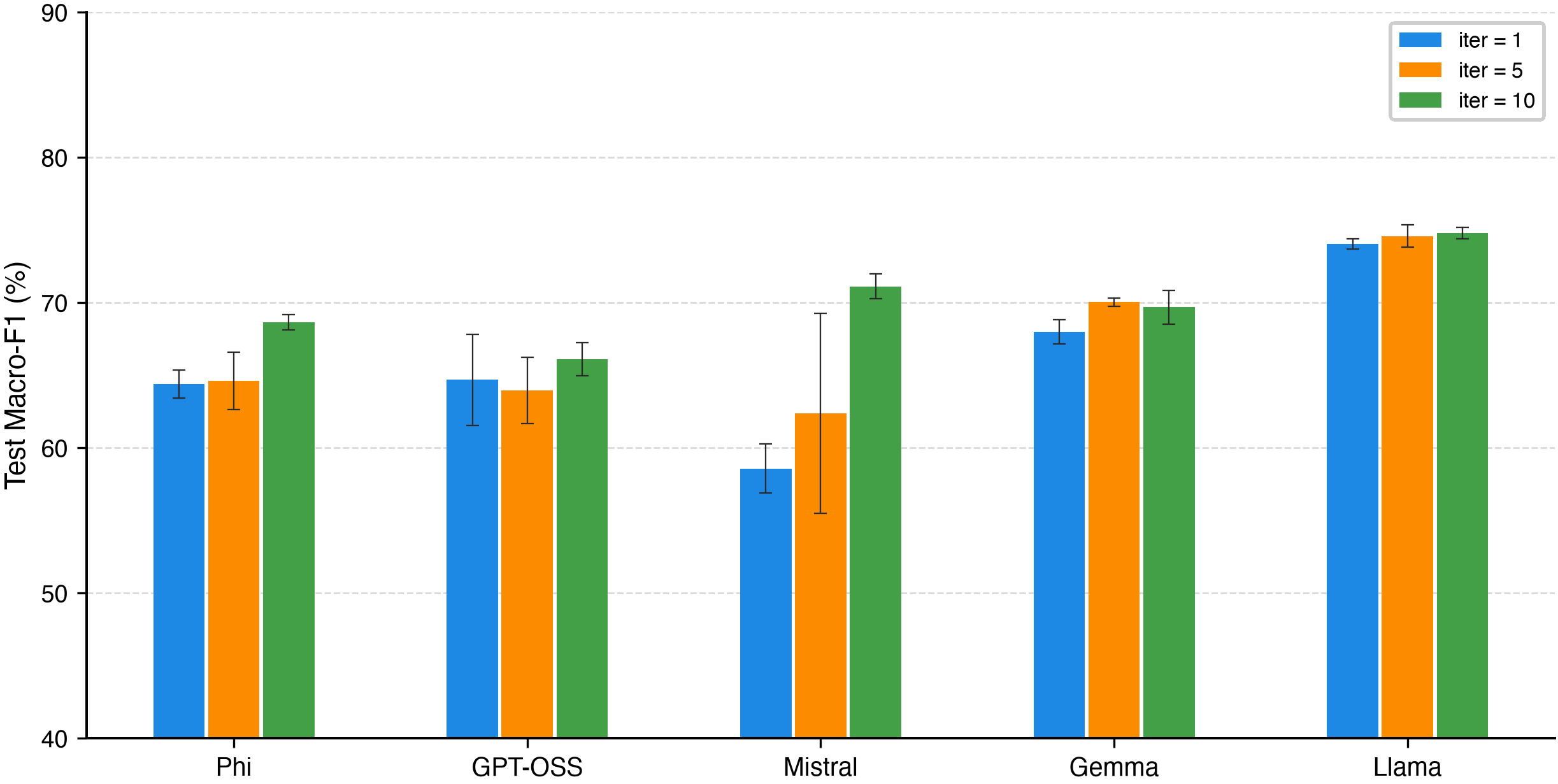}
    \caption{
    Effect of optimization budget on SemEval-2016 under the I+R+E
    setting. Bars show mean test Macro-F1 over three independent runs
    using 1, 5, and 10 optimization iterations; error bars denote
    standard deviation.
    }
    \label{fig:semeval-ire-iteration-budget}
\end{figure}

We examine the effect of optimization budget under the I+R+E setting
by varying the number of optimization iterations among 1, 5, and 10
on SemEval-2016. Figure~\ref{fig:semeval-ire-iteration-budget} reports
test Macro-F1 across all five backbone models.

Increasing the optimization budget from 1 to 10 iterations improves
mean Macro-F1 for all five backbones, although the magnitude of the
gain varies across models. The largest improvement is observed for
Mistral ($+12.55$ points), followed by Phi ($+4.23$), with smaller
gains for GPT-OSS, Gemma, and Llama. Paired two-sided $t$-tests across
the three runs show significant improvements from 1 to 10 iterations
for Mistral ($p=0.0018$), Gemma ($p=0.0210$), and Phi ($p=0.0211$),
but not for GPT-OSS or Llama. For 5 versus 10 iterations, no
statistically significant difference is detected for any backbone
with three runs ($p>0.05$).
\subsection{Implementation Details}
\label{app:impl_details}

All LLM calls are performed through remote APIs, with OpenRouter as the
primary provider. Stance classification uses a temperature of 0.2,
whereas reasoning-generation steps use a temperature of 0.7. For
few-shot evaluation, the final optimized I+R+E prompts are kept frozen
and augmented at inference time with training-set demonstrations
retrieved by embedding similarity. We evaluate $k=1$ and $k=3$
demonstrations per stance class.

\section{Additional Optimization Trajectories}
\label{app:trajectory_plots}

\begin{figure*}[!t]
    \centering
    \includegraphics[width=0.9\textwidth]{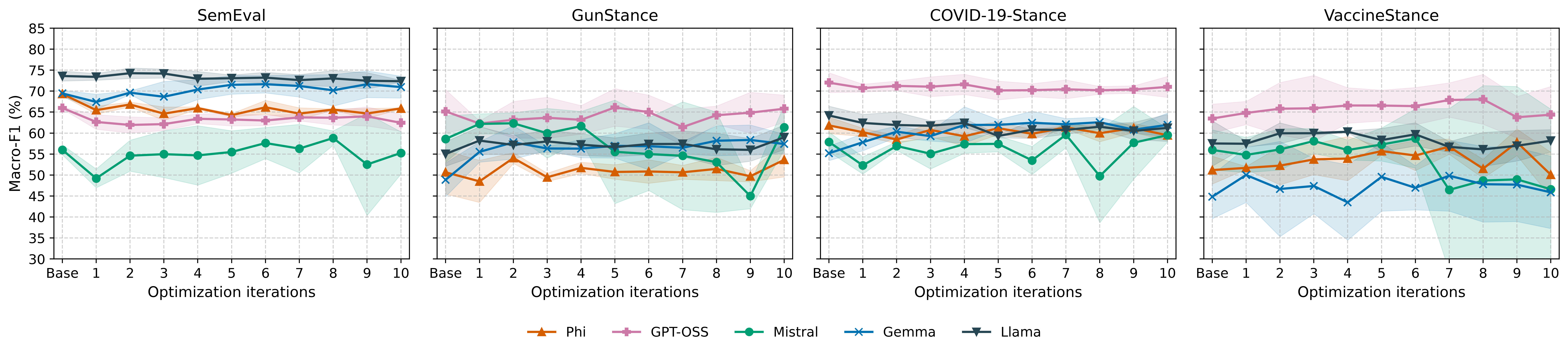}
    \caption{
    Validation Macro-F1 trajectories over 10 optimization iterations
    across the four datasets using the I strategy. Curves show the mean
    across three independent runs, and shaded regions denote standard
    deviation. ``Base'' denotes the validation Macro-F1 of the original
    prompt before optimization.
    }
    \label{fig:i_trajectory_appendix}
\end{figure*}

\begin{figure*}[!ht]
    \centering
    \includegraphics[width=0.9\textwidth]{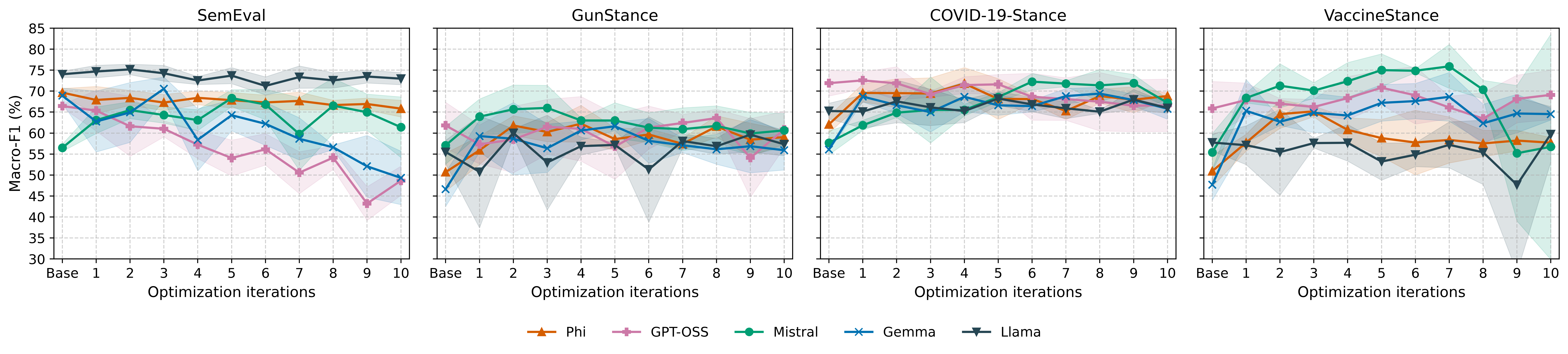}
    \caption{
    Validation Macro-F1 trajectories over 10 optimization iterations
    across the four datasets using the I+R strategy. Curves show the mean
    across three independent runs, and shaded regions denote standard
    deviation. ``Base'' denotes the validation Macro-F1 of the original
    prompt before optimization.
    }
    \label{fig:ir_trajectory_appendix}
\end{figure*}

Figures~\ref{fig:i_trajectory_appendix} and
\ref{fig:ir_trajectory_appendix} present the validation Macro-F1
trajectories for the I and I+R optimization strategies, respectively.
Each curve represents the mean validation Macro-F1 across three
independent runs, while the shaded regions indicate the corresponding
standard deviation. In both figures, ``Base'' denotes the validation
Macro-F1 of the original prompt before optimization.

\section{Detailed Dataset-Specific Results}
\label{app:detailed_results}

This section provides detailed target-level Macro-F1 results for the SemEval-2016, GunStance, and COVID-19-Stance datasets discussed in the main paper. Since VaccineStance contains a single target, its full results are reported in the main paper.

Table~\ref{tab:semeval-results-po} reports the detailed target-level results for SemEval-2016.
Table~\ref{tab:gunstance-results-po} reports the detailed target-level results for GunStance.
Table~\ref{tab:covid-19-stance-results-po} reports the detailed target-level results for COVID-19-Stance.

\begin{table*}[t]
\centering
\fontsize{7.8}{8.8}\selectfont
\setlength{\tabcolsep}{4pt}

{
\begin{tabular}{ll l l l l l}
\toprule
\multicolumn{2}{l}{}
& \multicolumn{5}{c}{\textbf{Model}} \\
\cmidrule(lr){3-7}
\textbf{Target}
& \textbf{Method}
& \textbf{Phi}
& \textbf{GPT-OSS}
& \textbf{Mistral}
& \textbf{Gemma}
& \textbf{Llama} \\
\midrule


\multirow{12}{*}{Atheism}

& Base
& 37.88 $\pm$ 1.41
& 42.71 $\pm$ 0.96
& 20.15 $\pm$ 0.98
& 60.48 $\pm$ 0.22
& 63.74 $\pm$ 1.36 \\

& Base+CoT
& \cellcolor{pos10}64.11 $\pm$ 0.56
& \cellcolor{neg7}33.65 $\pm$ 1.53
& \cellcolor{pos10}55.71 $\pm$ 0.50
& \cellcolor{pos4}63.34 $\pm$ 0.74
& \cellcolor{neg3}64.09 $\pm$ 0.34 \\

& Base+FS
& \cellcolor{pos7}45.92 $\pm$ 2.87
& \cellcolor{pos8}52.47 $\pm$ 1.98
& \cellcolor{pos10}51.08 $\pm$ 0.97
& \cellcolor{pos6}65.84 $\pm$ 0.03
& \cellcolor{pos7}\textbf{72.25 $\pm$ 0.23} \\

& Base+CoT+FS
& \cellcolor{pos10}66.17 $\pm$ 0.57
& \cellcolor{pos4}44.94 $\pm$ 1.62
& \cellcolor{pos10}59.68 $\pm$ 1.61
& \cellcolor{pos3}61.99 $\pm$ 0.04
& \cellcolor{pos3}68.84 $\pm$ 0.56 \\

& Base+CoS
& \cellcolor{pos9}51.23 $\pm$ 1.34
& \cellcolor{pos9}55.08 $\pm$ 3.22
& \cellcolor{pos10}56.77 $\pm$ 2.57
& \cellcolor{neg3}58.10 $\pm$ 0.93
& \cellcolor{pos1}64.56 $\pm$ 1.80 \\

& OPRO
& \cellcolor{pos10}62.97 $\pm$ 3.37
& \cellcolor{pos10}59.13 $\pm$ 1.74
& \cellcolor{pos7}28.76 $\pm$ 10.81
& \cellcolor{neg5}56.10 $\pm$ 5.85
& \cellcolor{neg2}62.32 $\pm$ 1.87 \\

& CAPO
& \cellcolor{pos8}47.49 $\pm$ 5.59
& \cellcolor{pos10}61.00 $\pm$ 5.34
& \cellcolor{pos9}33.23 $\pm$ 2.05
& \cellcolor{pos4}64.18 $\pm$ 3.10
& \cellcolor{pos6}69.33 $\pm$ 0.76 \\

& EvoPrompt(DE)
& \cellcolor{pos10}59.11 $\pm$ 4.07
& \cellcolor{pos10}58.20 $\pm$ 3.46
& \cellcolor{pos10}51.13 $\pm$ 5.33
& \cellcolor{pos4}63.80 $\pm$ 0.37
& \cellcolor{pos3}66.20 $\pm$ 0.88 \\

& EvoPrompt(GA)
& \cellcolor{pos10}60.68 $\pm$ 6.53
& \cellcolor{pos9}56.82 $\pm$ 6.42
& \cellcolor{pos10}47.79 $\pm$ 15.26
& \cellcolor{pos6}65.91 $\pm$ 3.39
& \cellcolor{neg2}61.76 $\pm$ 3.68 \\

& I
& \cellcolor{pos5}42.63 $\pm$ 0.68
& \cellcolor{neg2}41.06 $\pm$ 2.80
& \cellcolor{pos10}40.34 $\pm$ 16.36
& \cellcolor{pos1}61.01 $\pm$ 2.15
& \cellcolor{neg4}60.34 $\pm$ 2.43 \\

& I+R
& \cellcolor{pos10}63.09 $\pm$ 4.86
& \cellcolor{pos10}66.31 $\pm$ 4.80
& \cellcolor{pos10}60.31 $\pm$ 3.77
& \cellcolor{pos5}64.94 $\pm$ 5.47
& \cellcolor{pos7}71.67 $\pm$ 0.63 \\

& I+R+E
& \cellcolor{pos10}60.77 $\pm$ 6.30
& \cellcolor{pos10}70.64 $\pm$ 2.34
& \cellcolor{pos10}67.46 $\pm$ 2.35
& \cellcolor{pos7}67.56 $\pm$ 2.42
& \cellcolor{pos4}66.92 $\pm$ 1.30 \\

\midrule


\multirow{12}{*}{\shortstack{Climate change is \\ a real concern}}

& Base
& 45.44 $\pm$ 2.30
& 63.98 $\pm$ 0.86
& 58.42 $\pm$ 0.71
& 55.43 $\pm$ 0.26
& 76.53 $\pm$ 0.44 \\

& Base+CoT
& \cellcolor{pos10}66.55 $\pm$ 2.20
& \cellcolor{neg2}60.91 $\pm$ 2.81
& \cellcolor{pos7}67.07 $\pm$ 0.76
& \cellcolor{pos8}65.10 $\pm$ 0.08
& \cellcolor{pos3}78.18 $\pm$ 0.11 \\

& Base+FS
& \cellcolor{pos6}51.51 $\pm$ 2.42
& \cellcolor{pos2}65.38 $\pm$ 3.35
& \cellcolor{pos7}66.59 $\pm$ 1.09
& \cellcolor{pos8}64.59 $\pm$ 0.15
& \cellcolor{neg4}73.29 $\pm$ 0.48 \\

& Base+CoT+FS
& \cellcolor{pos9}53.94 $\pm$ 0.76
& \cellcolor{neg1}62.50 $\pm$ 1.20
& \cellcolor{pos6}64.12 $\pm$ 0.64
& \cellcolor{pos8}64.91 $\pm$ 0.10
& \cellcolor{pos2}77.35 $\pm$ 0.96 \\

& Base+CoS
& \cellcolor{pos10}61.03 $\pm$ 3.49
& \cellcolor{pos4}67.51 $\pm$ 0.41
& \cellcolor{pos8}69.03 $\pm$ 3.49
& \cellcolor{pos9}68.91 $\pm$ 7.52
& \cellcolor{neg6}70.48 $\pm$ 2.88 \\

& OPRO
& \cellcolor{pos9}59.22 $\pm$ 0.97
& \cellcolor{pos6}70.61 $\pm$ 3.02
& \cellcolor{pos2}59.61 $\pm$ 0.66
& \cellcolor{neg1}54.97 $\pm$ 0.53
& \cellcolor{pos1}76.72 $\pm$ 0.88 \\

& CAPO
& \cellcolor{pos10}62.13 $\pm$ 1.90
& \cellcolor{pos5}68.65 $\pm$ 2.40
& \cellcolor{pos6}65.01 $\pm$ 0.63
& \cellcolor{pos8}64.67 $\pm$ 5.03
& \cellcolor{pos2}78.49 $\pm$ 1.94 \\

& EvoPrompt(DE)
& \cellcolor{pos10}64.81 $\pm$ 1.50
& \cellcolor{pos3}66.24 $\pm$ 4.11
& \cellcolor{pos8}67.60 $\pm$ 14.32
& \cellcolor{pos8}66.06 $\pm$ 2.95
& \cellcolor{pos3}79.30 $\pm$ 5.44 \\

& EvoPrompt(GA)
& \cellcolor{pos10}63.67 $\pm$ 2.31
& \cellcolor{pos3}66.42 $\pm$ 2.28
& \cellcolor{pos5}62.98 $\pm$ 2.07
& \cellcolor{pos8}66.49 $\pm$ 2.86
& \cellcolor{neg1}75.87 $\pm$ 9.80 \\

& I
& \cellcolor{pos9}58.58 $\pm$ 1.86
& \cellcolor{neg3}61.34 $\pm$ 3.07
& \cellcolor{pos2}59.67 $\pm$ 9.46
& \cellcolor{pos6}62.18 $\pm$ 4.07
& \cellcolor{pos4}79.84 $\pm$ 0.26 \\

& I+R
& \cellcolor{pos10}75.43 $\pm$ 4.06
& \cellcolor{neg7}55.90 $\pm$ 2.15
& \cellcolor{pos8}68.46 $\pm$ 4.32
& \cellcolor{pos10}73.08 $\pm$ 3.64
& \cellcolor{pos2}77.56 $\pm$ 0.60 \\

& I+R+E
& \cellcolor{pos10}67.25 $\pm$ 3.62
& \cellcolor{neg3}61.64 $\pm$ 5.86
& \cellcolor{pos9}73.40 $\pm$ 1.77
& \cellcolor{pos8}66.00 $\pm$ 3.95
& \cellcolor{pos6}\textbf{81.79 $\pm$ 0.67} \\

\midrule


\multirow{12}{*}{Feminist movement}

& Base
& 59.82 $\pm$ 1.68
& 61.78 $\pm$ 1.86
& 58.37 $\pm$ 0.98
& 67.69 $\pm$ 0.02
& 72.94 $\pm$ 0.50 \\

& Base+CoT
& \cellcolor{pos3}63.95 $\pm$ 1.70
& \cellcolor{pos1}61.80 $\pm$ 0.43
& \cellcolor{pos7}68.83 $\pm$ 0.94
& \cellcolor{neg2}65.50 $\pm$ 0.36
& \cellcolor{neg3}68.96 $\pm$ 0.97 \\

& Base+FS
& \cellcolor{pos2}61.45 $\pm$ 2.47
& \cellcolor{neg1}61.52 $\pm$ 1.69
& \cellcolor{pos5}62.46 $\pm$ 0.47
& \cellcolor{pos2}68.78 $\pm$ 0.20
& \cellcolor{neg5}68.24 $\pm$ 0.68 \\

& Base+CoT+FS
& \cellcolor{pos3}64.39 $\pm$ 2.26
& \cellcolor{pos2}63.22 $\pm$ 1.84
& \cellcolor{pos8}64.26 $\pm$ 0.34
& \cellcolor{neg2}65.49 $\pm$ 0.14
& \cellcolor{neg3}69.47 $\pm$ 0.26 \\

& Base+CoS
& \cellcolor{neg8}50.08 $\pm$ 3.94
& \cellcolor{neg7}53.26 $\pm$ 1.61
& \cellcolor{pos3}60.45 $\pm$ 0.06
& \cellcolor{neg8}56.20 $\pm$ 2.38
& \cellcolor{neg10}55.91 $\pm$ 2.21 \\

& OPRO
& \cellcolor{pos7}67.06 $\pm$ 3.93
& \cellcolor{pos6}68.22 $\pm$ 2.34
& \cellcolor{pos1}59.33 $\pm$ 4.39
& \cellcolor{neg1}67.45 $\pm$ 0.19
& \cellcolor{pos2}74.63 $\pm$ 1.16 \\

& CAPO
& \cellcolor{pos5}64.67 $\pm$ 1.26
& \cellcolor{pos7}68.99 $\pm$ 0.79
& \cellcolor{pos6}64.53 $\pm$ 0.36
& \cellcolor{neg2}65.80 $\pm$ 2.72
& \cellcolor{pos2}74.14 $\pm$ 1.70 \\

& EvoPrompt(DE)
& \cellcolor{pos8}68.96 $\pm$ 0.49
& \cellcolor{pos5}66.26 $\pm$ 1.19
& \cellcolor{pos8}67.66 $\pm$ 1.51
& \cellcolor{neg1}67.32 $\pm$ 3.63
& \cellcolor{pos3}\textbf{75.00 $\pm$ 3.02} \\

& EvoPrompt(GA)
& \cellcolor{pos7}67.48 $\pm$ 1.53
& \cellcolor{pos4}65.59 $\pm$ 1.94
& \cellcolor{pos3}60.59 $\pm$ 3.77
& \cellcolor{neg4}64.54 $\pm$ 0.31
& \cellcolor{pos1}73.11 $\pm$ 3.72 \\

& I
& \cellcolor{pos5}63.84 $\pm$ 2.68
& \cellcolor{pos1}62.27 $\pm$ 1.01
& \cellcolor{pos6}64.38 $\pm$ 5.64
& \cellcolor{pos3}70.59 $\pm$ 3.27
& \cellcolor{pos2}74.81 $\pm$ 0.06 \\

& I+R
& \cellcolor{pos4}63.15 $\pm$ 1.71
& \cellcolor{neg2}60.44 $\pm$ 1.97
& \cellcolor{pos4}62.29 $\pm$ 3.01
& \cellcolor{neg1}67.13 $\pm$ 4.16
& \cellcolor{pos1}73.64 $\pm$ 1.40 \\

& I+R+E
& \cellcolor{pos7}67.11 $\pm$ 1.30
& \cellcolor{pos3}64.69 $\pm$ 0.88
& \cellcolor{pos8}68.20 $\pm$ 1.36
& \cellcolor{pos1}67.86 $\pm$ 2.36
& \cellcolor{pos1}73.43 $\pm$ 0.59 \\

\midrule


\multirow{12}{*}{Hillary Clinton}

& Base
& 73.31 $\pm$ 0.31
& 74.59 $\pm$ 1.59
& 73.86 $\pm$ 0.54
& 80.94 $\pm$ 0.21
& 83.81 $\pm$ 0.51 \\

& Base+CoT
& \cellcolor{pos2}74.42 $\pm$ 3.97
& \cellcolor{pos1}74.35 $\pm$ 2.21
& \cellcolor{pos1}81.25 $\pm$ 0.20
& \cellcolor{neg2}78.66 $\pm$ 0.18
& \cellcolor{neg1}82.27 $\pm$ 0.46 \\

& Base+FS
& \cellcolor{neg6}66.52 $\pm$ 2.00
& \cellcolor{pos1}75.34 $\pm$ 0.58
& \cellcolor{pos7}81.07 $\pm$ 1.11
& \cellcolor{neg1}80.25 $\pm$ 0.49
& \cellcolor{neg2}82.46 $\pm$ 0.43 \\

& Base+CoT+FS
& \cellcolor{pos6}78.68 $\pm$ 3.02
& \cellcolor{pos3}76.05 $\pm$ 1.00
& \cellcolor{pos7}81.59 $\pm$ 0.86
& \cellcolor{neg3}77.14 $\pm$ 0.30
& \cellcolor{neg4}78.96 $\pm$ 0.20 \\

& Base+CoS
& \cellcolor{neg10}56.69 $\pm$ 0.46
& \cellcolor{neg7}67.36 $\pm$ 1.61
& \cellcolor{pos2}75.12 $\pm$ 0.71
& \cellcolor{neg10}58.72 $\pm$ 0.73
& \cellcolor{neg10}64.53 $\pm$ 3.34 \\

& OPRO
& \cellcolor{pos7}81.56 $\pm$ 0.42
& \cellcolor{pos1}75.50 $\pm$ 1.78
& \cellcolor{pos3}75.91 $\pm$ 2.28
& \cellcolor{neg1}80.78 $\pm$ 0.41
& \cellcolor{pos1}84.05 $\pm$ 0.39 \\

& CAPO
& \cellcolor{pos6}80.10 $\pm$ 2.22
& \cellcolor{pos5}79.45 $\pm$ 0.81
& \cellcolor{pos5}78.01 $\pm$ 0.39
& \cellcolor{neg3}78.50 $\pm$ 4.65
& \cellcolor{neg1}83.14 $\pm$ 1.46 \\

& EvoPrompt(DE)
& \cellcolor{pos8}83.03 $\pm$ 0.10
& \cellcolor{pos4}77.90 $\pm$ 2.05
& \cellcolor{pos5}78.58 $\pm$ 2.21
& \cellcolor{neg2}79.40 $\pm$ 0.42
& \cellcolor{neg1}83.36 $\pm$ 0.61 \\

& EvoPrompt(GA)
& \cellcolor{pos7}80.88 $\pm$ 0.67
& \cellcolor{pos3}77.54 $\pm$ 2.69
& \cellcolor{pos6}79.32 $\pm$ 2.31
& \cellcolor{pos2}81.98 $\pm$ 2.06
& \cellcolor{neg2}82.63 $\pm$ 2.38 \\

& I
& \cellcolor{pos6}80.13 $\pm$ 2.27
& \cellcolor{pos1}75.49 $\pm$ 0.54
& \cellcolor{pos8}83.67 $\pm$ 2.47
& \cellcolor{neg1}80.77 $\pm$ 0.67
& \cellcolor{pos1}84.13 $\pm$ 0.41 \\

& I+R
& \cellcolor{pos7}80.47 $\pm$ 1.62
& \cellcolor{pos1}75.34 $\pm$ 0.78
& \cellcolor{pos7}82.58 $\pm$ 1.97
& \cellcolor{neg1}80.82 $\pm$ 1.01
& \cellcolor{pos1}84.47 $\pm$ 0.72 \\

& I+R+E
& \cellcolor{pos8}82.65 $\pm$ 1.74
& \cellcolor{pos2}75.71 $\pm$ 1.80
& \cellcolor{pos8}83.93 $\pm$ 1.21
& \cellcolor{neg2}79.75 $\pm$ 1.79
& \cellcolor{pos1}\textbf{84.68 $\pm$ 0.55} \\

\midrule


\multirow{12}{*}{\shortstack{Legalization of \\ abortion}}

& Base
& 60.39 $\pm$ 0.74
& 57.05 $\pm$ 1.19
& 58.71 $\pm$ 0.57
& 64.55 $\pm$ 1.00
& 70.58 $\pm$ 0.73 \\

& Base+CoT
& \cellcolor{pos3}62.10 $\pm$ 1.39
& \cellcolor{pos1}57.02 $\pm$ 1.15
& \cellcolor{pos5}62.44 $\pm$ 0.68
& \cellcolor{pos3}66.52 $\pm$ 1.35
& \cellcolor{neg5}65.16 $\pm$ 0.94 \\

& Base+FS
& \cellcolor{pos4}64.12 $\pm$ 2.10
& \cellcolor{neg1}56.16 $\pm$ 1.31
& \cellcolor{pos4}62.24 $\pm$ 0.80
& \cellcolor{pos3}67.05 $\pm$ 0.31
& \cellcolor{neg6}65.16 $\pm$ 0.87 \\

& Base+CoT+FS
& \cellcolor{pos4}63.06 $\pm$ 4.10
& \cellcolor{pos1}56.88 $\pm$ 1.20
& \cellcolor{pos8}66.93 $\pm$ 0.15
& \cellcolor{pos1}64.13 $\pm$ 0.32
& \cellcolor{neg7}61.56 $\pm$ 0.54 \\

& Base+CoS
& \cellcolor{neg9}48.00 $\pm$ 4.50
& \cellcolor{pos1}57.71 $\pm$ 1.48
& \cellcolor{neg1}58.31 $\pm$ 0.18
& \cellcolor{neg6}57.66 $\pm$ 0.47
& \cellcolor{neg8}59.33 $\pm$ 0.61 \\

& OPRO
& \cellcolor{pos6}66.65 $\pm$ 0.59
& \cellcolor{pos3}59.47 $\pm$ 2.32
& \cellcolor{pos1}59.44 $\pm$ 0.76
& \cellcolor{neg1}64.51 $\pm$ 1.53
& \cellcolor{neg2}69.53 $\pm$ 0.16 \\

& CAPO
& \cellcolor{pos4}64.15 $\pm$ 2.14
& \cellcolor{pos2}58.51 $\pm$ 1.20
& \cellcolor{pos3}60.83 $\pm$ 0.21
& \cellcolor{pos2}65.84 $\pm$ 1.65
& \cellcolor{neg2}69.23 $\pm$ 0.68 \\

& EvoPrompt(DE)
& \cellcolor{pos4}64.18 $\pm$ 0.13
& \cellcolor{pos3}59.36 $\pm$ 0.57
& \cellcolor{pos4}61.78 $\pm$ 0.91
& \cellcolor{neg2}63.23 $\pm$ 0.99
& \cellcolor{neg1}70.28 $\pm$ 2.47 \\

& EvoPrompt(GA)
& \cellcolor{pos4}63.71 $\pm$ 2.53
& \cellcolor{pos1}57.86 $\pm$ 3.37
& \cellcolor{pos3}60.93 $\pm$ 0.62
& \cellcolor{pos3}66.63 $\pm$ 1.40
& \cellcolor{neg2}69.16 $\pm$ 1.18 \\

& I
& \cellcolor{pos3}62.94 $\pm$ 2.74
& \cellcolor{pos1}57.92 $\pm$ 1.93
& \cellcolor{pos3}61.59 $\pm$ 1.40
& \cellcolor{pos3}66.77 $\pm$ 2.33
& \cellcolor{neg3}67.99 $\pm$ 1.27 \\

& I+R
& \cellcolor{pos6}66.11 $\pm$ 0.51
& \cellcolor{neg1}56.81 $\pm$ 1.67
& \cellcolor{pos6}65.61 $\pm$ 1.61
& \cellcolor{neg3}62.22 $\pm$ 6.23
& \cellcolor{neg4}66.89 $\pm$ 0.38 \\

& I+R+E
& \cellcolor{pos5}65.36 $\pm$ 0.75
& \cellcolor{pos1}57.84 $\pm$ 1.53
& \cellcolor{pos4}62.62 $\pm$ 0.47
& \cellcolor{pos3}67.20 $\pm$ 1.44
& \cellcolor{neg4}67.10 $\pm$ 0.30 \\

\bottomrule
\end{tabular}
}

\caption{
Per-target results on SemEval-2016 using Macro-F1.
I = instruction-only; I+R = instruction + reasoning; I+R+E = instruction
+ per-example reasoning. Base = original prompt; CoT = chain-of-thought
prompting; FS = few-shot prompting; CoS = Chain-of-Stance prompting.
Values are mean $\pm$ standard deviation across three runs.
Shaded cells indicate differences from the corresponding Base score:
gray denotes improvements, red denotes decreases, and darker shades indicate
larger absolute differences. The best score for each target across all
models and methods is shown in \textbf{bold}.
}
\label{tab:semeval-results-po}

\end{table*}

\begin{table*}[t]
\centering
\fontsize{7.8}{8.8}\selectfont

{
\begin{tabular}{l l l l l l l}
\toprule
\textbf{Target}
& \textbf{Method}
& \textbf{Phi}
& \textbf{GPT-OSS}
& \textbf{Mistral}
& \textbf{Gemma}
& \textbf{Llama} \\
\midrule


\multirow{12}{*}{\shortstack{Guns should \\ be banned}}

& Base
& 39.84 $\pm$ 1.92
& 57.53 $\pm$ 0.81
& 53.75 $\pm$ 0.56
& 43.49 $\pm$ 0.23
& 52.17 $\pm$ 0.50 \\

& Base+CoT
& \cellcolor{pos8}60.73 $\pm$ 1.24
& \cellcolor{neg3}57.26 $\pm$ 1.20
& \cellcolor{neg1}54.44 $\pm$ 0.43
& \cellcolor{pos4}49.46 $\pm$ 0.08
& \cellcolor{pos3}54.66 $\pm$ 0.81 \\

& Base+FS
& \cellcolor{pos9}51.99 $\pm$ 2.07
& \cellcolor{pos2}58.69 $\pm$ 0.70
& \cellcolor{pos8}62.86 $\pm$ 0.13
& \cellcolor{pos9}57.18 $\pm$ 0.14
& \cellcolor{pos2}53.61 $\pm$ 0.29 \\

& Base+CoT+FS
& \cellcolor{pos8}59.92 $\pm$ 0.53
& \cellcolor{neg4}56.41 $\pm$ 0.76
& \cellcolor{pos4}58.77 $\pm$ 0.58
& \cellcolor{pos9}62.44 $\pm$ 0.08
& \cellcolor{pos6}55.84 $\pm$ 1.10 \\

& Base+CoS
& \cellcolor{pos3}42.52 $\pm$ 1.06
& \cellcolor{neg7}50.39 $\pm$ 1.50
& \cellcolor{pos7}62.19 $\pm$ 1.11
& \cellcolor{pos9}57.98 $\pm$ 0.49
& \cellcolor{neg6}46.87 $\pm$ 0.63 \\

& OPRO
& \cellcolor{pos9}54.13 $\pm$ 3.03
& \cellcolor{neg1}56.74 $\pm$ 3.02
& \cellcolor{pos1}54.09 $\pm$ 0.62
& \cellcolor{neg1}42.60 $\pm$ 0.71
& \cellcolor{neg1}52.14 $\pm$ 0.84 \\

& CAPO
& \cellcolor{pos9}54.64 $\pm$ 1.57
& \cellcolor{pos2}58.79 $\pm$ 1.18
& \cellcolor{pos6}60.73 $\pm$ 2.23
& \cellcolor{pos6}49.45 $\pm$ 2.95
& \cellcolor{pos1}52.96 $\pm$ 3.35 \\

& EvoPrompt(DE)
& \cellcolor{pos10}55.32 $\pm$ 3.46
& \cellcolor{neg2}55.76 $\pm$ 0.16
& \cellcolor{pos2}55.28 $\pm$ 2.28
& \cellcolor{pos6}49.04 $\pm$ 6.53
& \cellcolor{pos6}57.56 $\pm$ 1.13 \\

& EvoPrompt(GA)
& \cellcolor{pos10}55.49 $\pm$ 3.44
& \cellcolor{pos1}58.05 $\pm$ 1.99
& \cellcolor{pos3}56.10 $\pm$ 1.78
& \cellcolor{pos8}52.83 $\pm$ 4.08
& \cellcolor{neg1}51.82 $\pm$ 3.33 \\

& I
& \cellcolor{pos8}50.80 $\pm$ 2.01
& \cellcolor{pos3}60.34 $\pm$ 0.48
& \cellcolor{pos6}60.10 $\pm$ 5.70
& \cellcolor{pos9}56.14 $\pm$ 3.19
& \cellcolor{pos3}54.71 $\pm$ 1.46 \\

& I+R
& \cellcolor{pos10}57.96 $\pm$ 5.38
& \cellcolor{pos3}60.51 $\pm$ 1.93
& \cellcolor{pos8}62.82 $\pm$ 3.24
& \cellcolor{pos10}61.81 $\pm$ 1.56
& \cellcolor{pos5}57.00 $\pm$ 3.56 \\

& I+R+E
& \cellcolor{pos10}58.02 $\pm$ 1.75
& \cellcolor{pos3}60.07 $\pm$ 3.16
& \cellcolor{pos8}64.78 $\pm$ 1.89
& \cellcolor{pos9}57.50 $\pm$ 4.79
& \cellcolor{pos5}56.58 $\pm$ 4.40 \\

\midrule


\multirow{12}{*}{\shortstack{Guns should \\ be regulated}}

& Base
& 39.27 $\pm$ 1.26
& 63.16 $\pm$ 0.74
& 58.73 $\pm$ 0.50
& 50.69 $\pm$ 0.38
& 57.54 $\pm$ 0.44 \\

& Base+CoT
& \cellcolor{pos10}61.62 $\pm$ 0.68
& \cellcolor{pos1}62.27 $\pm$ 0.30
& \cellcolor{pos3}60.65 $\pm$ 0.71
& \cellcolor{pos6}55.49 $\pm$ 0.06
& \cellcolor{pos2}57.88 $\pm$ 0.47 \\

& Base+FS
& \cellcolor{pos8}50.95 $\pm$ 1.67
& \cellcolor{neg1}62.50 $\pm$ 0.53
& \cellcolor{pos6}65.40 $\pm$ 0.42
& \cellcolor{pos7}58.88 $\pm$ 0.13
& \cellcolor{pos6}62.63 $\pm$ 0.24 \\

& Base+CoT+FS
& \cellcolor{pos10}58.71 $\pm$ 0.66
& \cellcolor{neg1}61.29 $\pm$ 0.98
& \cellcolor{pos6}64.66 $\pm$ 0.34
& \cellcolor{pos9}64.68 $\pm$ 0.73
& \cellcolor{pos5}62.19 $\pm$ 0.91 \\

& Base+CoS
& \cellcolor{pos5}43.81 $\pm$ 0.55
& \cellcolor{neg9}50.02 $\pm$ 1.04
& \cellcolor{pos5}62.90 $\pm$ 1.45
& \cellcolor{pos8}60.61 $\pm$ 2.69
& \cellcolor{neg5}52.60 $\pm$ 0.14 \\

& OPRO
& \cellcolor{pos10}57.66 $\pm$ 2.55
& \cellcolor{neg2}61.46 $\pm$ 1.12
& \cellcolor{pos2}60.64 $\pm$ 4.46
& \cellcolor{neg3}48.65 $\pm$ 1.60
& \cellcolor{neg2}56.42 $\pm$ 1.29 \\

& CAPO
& \cellcolor{pos10}58.73 $\pm$ 3.21
& \cellcolor{neg2}62.15 $\pm$ 0.36
& \cellcolor{pos5}63.65 $\pm$ 0.58
& \cellcolor{pos7}58.49 $\pm$ 1.21
& \cellcolor{neg2}55.80 $\pm$ 0.78 \\

& EvoPrompt(DE)
& \cellcolor{pos10}55.51 $\pm$ 3.87
& \cellcolor{neg3}61.06 $\pm$ 1.87
& \cellcolor{pos5}63.52 $\pm$ 1.44
& \cellcolor{pos5}55.36 $\pm$ 2.39
& \cellcolor{neg1}56.54 $\pm$ 0.91 \\

& EvoPrompt(GA)
& \cellcolor{pos10}56.79 $\pm$ 0.86
& \cellcolor{neg3}61.03 $\pm$ 1.34
& \cellcolor{pos4}62.17 $\pm$ 0.20
& \cellcolor{pos6}56.93 $\pm$ 2.21
& \cellcolor{neg2}55.85 $\pm$ 1.26 \\

& I
& \cellcolor{pos10}56.57 $\pm$ 2.82
& \cellcolor{pos1}63.31 $\pm$ 1.77
& \cellcolor{pos5}62.76 $\pm$ 2.67
& \cellcolor{pos8}60.91 $\pm$ 0.12
& \cellcolor{pos3}60.38 $\pm$ 1.14 \\

& I+R
& \cellcolor{pos10}64.01 $\pm$ 1.78
& \cellcolor{neg2}61.63 $\pm$ 0.56
& \cellcolor{pos6}64.53 $\pm$ 0.70
& \cellcolor{pos10}66.29 $\pm$ 0.74
& \cellcolor{pos4}60.97 $\pm$ 0.94 \\

& I+R+E
& \cellcolor{pos10}61.47 $\pm$ 1.79
& \cellcolor{pos2}64.16 $\pm$ 0.76
& \cellcolor{pos6}65.24 $\pm$ 2.73
& \cellcolor{pos9}62.93 $\pm$ 2.94
& \cellcolor{pos4}61.48 $\pm$ 3.73 \\

\bottomrule
\end{tabular}
}

\caption{
Per-target results on GunStance using Macro-F1.
I = instruction-only; I+R = instruction + reasoning; I+R+E = instruction + per-example reasoning.
Base = original prompt; CoT = chain-of-thought prompting; FS = few-shot prompting;
CoS = Chain-of-Stance prompting.
Values are mean $\pm$ standard deviation across three runs.
Shaded cells indicate differences from the corresponding Base score within each model column:
gray denotes improvements, red denotes decreases, and darker shades indicate larger absolute differences.
The best score for each target across all models and methods is shown in \textbf{bold}.
}
\label{tab:gunstance-results-po}

\end{table*}

\begin{table*}[t]
\centering
\fontsize{7.5}{8.5}\selectfont

{
\begin{tabular}{ll l l l l l}
\toprule
\textbf{Target}
& \textbf{Method}
& \textbf{Phi}
& \textbf{GPT-OSS}
& \textbf{Mistral}
& \textbf{Gemma}
& \textbf{Llama} \\
\midrule


\multirow{12}{*}{Anthony Fauci}

& Base
& 57.47 $\pm$ 1.84
& 70.76 $\pm$ 3.03
& 72.65 $\pm$ 0.13
& 58.08 $\pm$ 0.43
& 70.91 $\pm$ 0.33 \\

& Base+CoT
& \cellcolor{pos8}66.75 $\pm$ 0.46
& \cellcolor{pos4}73.35 $\pm$ 1.36
& \cellcolor{pos3}73.71 $\pm$ 0.83
& \cellcolor{pos5}61.95 $\pm$ 1.35
& \cellcolor{neg1}69.71 $\pm$ 0.42 \\

& Base+FS
& \cellcolor{pos6}63.23 $\pm$ 2.49
& \cellcolor{pos3}73.28 $\pm$ 0.96
& \cellcolor{pos3}75.20 $\pm$ 0.77
& \cellcolor{pos6}64.53 $\pm$ 0.34
& \cellcolor{pos3}73.28 $\pm$ 0.50 \\

& Base+CoT+FS
& \cellcolor{pos6}63.29 $\pm$ 0.86
& \cellcolor{pos6}74.83 $\pm$ 0.84
& \cellcolor{neg2}70.62 $\pm$ 0.63
& \cellcolor{pos8}67.22 $\pm$ 0.54
& \cellcolor{pos2}71.31 $\pm$ 1.10 \\

& Base+CoS
& \cellcolor{neg4}53.62 $\pm$ 2.26
& \cellcolor{neg10}52.93 $\pm$ 0.69
& \cellcolor{neg8}62.66 $\pm$ 1.01
& \cellcolor{neg1}57.88 $\pm$ 3.59
& \cellcolor{neg8}60.67 $\pm$ 3.26 \\

& OPRO
& \cellcolor{pos9}70.12 $\pm$ 0.32
& \cellcolor{neg2}69.48 $\pm$ 2.32
& \cellcolor{neg1}72.61 $\pm$ 2.42
& \cellcolor{pos2}59.12 $\pm$ 0.50
& \cellcolor{neg2}69.39 $\pm$ 0.49 \\

& CAPO
& \cellcolor{pos10}72.77 $\pm$ 0.44
& \cellcolor{pos3}73.30 $\pm$ 1.53
& \cellcolor{neg3}70.45 $\pm$ 2.74
& \cellcolor{pos1}58.74 $\pm$ 1.44
& \cellcolor{neg2}69.17 $\pm$ 0.54 \\

& EvoPrompt(DE)
& \cellcolor{pos8}69.43 $\pm$ 1.34
& \cellcolor{pos3}72.81 $\pm$ 2.32
& \cellcolor{neg2}71.45 $\pm$ 0.49
& \cellcolor{pos6}63.48 $\pm$ 0.45
& \cellcolor{pos2}72.16 $\pm$ 2.74 \\

& EvoPrompt(GA)
& \cellcolor{pos10}73.01 $\pm$ 0.72
& \cellcolor{pos2}72.24 $\pm$ 1.88
& \cellcolor{pos1}72.87 $\pm$ 1.75
& \cellcolor{pos6}63.49 $\pm$ 1.65
& \cellcolor{pos3}73.66 $\pm$ 3.40 \\

& I
& \cellcolor{pos8}69.42 $\pm$ 1.22
& \cellcolor{pos5}75.42 $\pm$ 1.36
& \cellcolor{pos1}72.75 $\pm$ 1.19
& \cellcolor{pos2}59.96 $\pm$ 3.58
& \cellcolor{neg1}70.02 $\pm$ 0.33 \\

& I+R
& \cellcolor{pos10}74.75 $\pm$ 1.18
& \cellcolor{pos5}74.80 $\pm$ 2.10
& \cellcolor{pos6}77.76 $\pm$ 1.81
& \cellcolor{pos8}67.47 $\pm$ 7.64
& \cellcolor{neg1}69.91 $\pm$ 0.29 \\

& I+R+E
& \cellcolor{pos10}75.65 $\pm$ 3.43
& \cellcolor{pos3}72.77 $\pm$ 2.26
& \cellcolor{pos7}\textbf{80.62 $\pm$ 1.36}
& \cellcolor{pos4}61.79 $\pm$ 4.26
& \cellcolor{pos5}75.40 $\pm$ 2.44 \\

\midrule


\multirow{12}{*}{Face masks}

& Base
& 51.62 $\pm$ 2.71
& 75.62 $\pm$ 1.27
& 66.12 $\pm$ 0.73
& 67.78 $\pm$ 0.00
& 78.26 $\pm$ 0.09 \\

& Base+CoT
& \cellcolor{pos10}76.28 $\pm$ 3.36
& \cellcolor{pos2}75.89 $\pm$ 1.73
& \cellcolor{pos8}75.88 $\pm$ 0.88
& \cellcolor{pos5}71.37 $\pm$ 1.20
& \cellcolor{pos1}78.13 $\pm$ 1.00 \\
& Base+FS
& \cellcolor{pos4}54.71 $\pm$ 1.11
& \cellcolor{neg3}73.60 $\pm$ 0.67
& \cellcolor{pos7}73.59 $\pm$ 0.85
& \cellcolor{pos5}72.63 $\pm$ 0.92
& \cellcolor{neg4}75.04 $\pm$ 0.48 \\

& Base+CoT+FS
& \cellcolor{pos10}71.08 $\pm$ 2.20
& \cellcolor{neg3}71.82 $\pm$ 0.48
& \cellcolor{pos9}79.68 $\pm$ 1.20
& \cellcolor{pos6}73.45 $\pm$ 0.18
& \cellcolor{pos1}77.47 $\pm$ 0.30 \\

& Base+CoS
& \cellcolor{pos8}62.44 $\pm$ 3.54
& \cellcolor{neg7}68.43 $\pm$ 0.51
& \cellcolor{pos4}69.94 $\pm$ 3.85
& \cellcolor{pos2}68.81 $\pm$ 1.02
& \cellcolor{neg8}67.20 $\pm$ 0.52 \\

& OPRO
& \cellcolor{pos9}66.16 $\pm$ 4.32
& \cellcolor{neg4}72.57 $\pm$ 1.53
& \cellcolor{pos2}67.85 $\pm$ 0.08
& \cellcolor{neg4}64.03 $\pm$ 3.75
& \cellcolor{pos1}78.84 $\pm$ 0.54 \\

& CAPO
& \cellcolor{pos9}64.51 $\pm$ 2.35
& \cellcolor{pos2}77.00 $\pm$ 0.46
& \cellcolor{pos7}73.62 $\pm$ 1.93
& \cellcolor{neg2}66.63 $\pm$ 4.40
& \cellcolor{pos2}79.71 $\pm$ 2.49 \\

& EvoPrompt(DE)
& \cellcolor{pos10}67.36 $\pm$ 6.46
& \cellcolor{neg2}74.59 $\pm$ 3.18
& \cellcolor{pos6}71.34 $\pm$ 0.59
& \cellcolor{pos7}75.10 $\pm$ 0.57
& \cellcolor{pos2}80.16 $\pm$ 2.54 \\

& EvoPrompt(GA)
& \cellcolor{pos9}66.08 $\pm$ 3.90
& \cellcolor{neg1}75.27 $\pm$ 2.15
& \cellcolor{pos6}71.51 $\pm$ 1.81
& \cellcolor{pos6}73.43 $\pm$ 2.07
& \cellcolor{pos3}80.27 $\pm$ 1.71 \\

& I
& \cellcolor{pos9}66.40 $\pm$ 3.62
& \cellcolor{neg1}75.44 $\pm$ 1.46
& \cellcolor{pos7}75.04 $\pm$ 6.50
& \cellcolor{pos2}69.77 $\pm$ 3.67
& \cellcolor{neg2}76.45 $\pm$ 4.37 \\

& I+R
& \cellcolor{pos10}78.69 $\pm$ 1.74
& \cellcolor{neg2}73.91 $\pm$ 0.84
& \cellcolor{pos9}79.08 $\pm$ 2.26
& \cellcolor{pos7}76.02 $\pm$ 2.43
& \cellcolor{pos1}78.75 $\pm$ 3.19 \\

& I+R+E
& \cellcolor{pos10}70.89 $\pm$ 1.54
& \cellcolor{pos2}76.67 $\pm$ 2.30
& \cellcolor{pos9}80.34 $\pm$ 0.88
& \cellcolor{pos7}76.03 $\pm$ 1.48
& \cellcolor{pos4}\textbf{81.49 $\pm$ 1.90} \\

\midrule


\multirow{12}{*}{School closures}

& Base
& 33.87 $\pm$ 2.22
& 70.74 $\pm$ 1.77
& 40.97 $\pm$ 0.93
& 39.44 $\pm$ 0.31
& 40.45 $\pm$ 0.73 \\

& Base+CoT
& \cellcolor{pos10}50.85 $\pm$ 1.11
& \cellcolor{pos1}70.20 $\pm$ 1.70
& \cellcolor{pos10}59.90 $\pm$ 1.20
& \cellcolor{pos9}51.02 $\pm$ 0.28
& \cellcolor{pos10}60.60 $\pm$ 0.32 \\

& Base+FS
& \cellcolor{pos3}35.91 $\pm$ 2.51
& \cellcolor{pos2}71.77 $\pm$ 1.49
& \cellcolor{pos7}48.65 $\pm$ 0.73
& \cellcolor{pos9}51.44 $\pm$ 0.00
& \cellcolor{pos10}61.93 $\pm$ 0.91 \\

& Base+CoT+FS
& \cellcolor{pos8}44.34 $\pm$ 3.68
& \cellcolor{pos3}72.21 $\pm$ 1.52
& \cellcolor{pos10}58.17 $\pm$ 0.68
& \cellcolor{pos10}47.16 $\pm$ 0.87
& \cellcolor{pos10}61.10 $\pm$ 1.20 \\

& Base+CoS
& \cellcolor{pos6}39.31 $\pm$ 1.36
& \cellcolor{neg10}55.01 $\pm$ 0.21
& \cellcolor{pos3}43.67 $\pm$ 1.48
& \cellcolor{pos4}43.21 $\pm$ 4.74
& \cellcolor{neg6}34.36 $\pm$ 0.97 \\

& OPRO
& \cellcolor{pos8}45.76 $\pm$ 3.54
& \cellcolor{neg7}62.98 $\pm$ 1.06
& \cellcolor{neg6}34.50 $\pm$ 5.96
& \cellcolor{neg2}37.60 $\pm$ 2.42
& \cellcolor{pos2}41.67 $\pm$ 3.48 \\

& CAPO
& \cellcolor{pos7}42.39 $\pm$ 2.61
& \cellcolor{neg1}69.84 $\pm$ 1.31
& \cellcolor{pos2}42.57 $\pm$ 1.90
& \cellcolor{pos8}50.63 $\pm$ 1.96
& \cellcolor{pos3}43.29 $\pm$ 1.69 \\

& EvoPrompt(DE)
& \cellcolor{pos8}44.59 $\pm$ 3.22
& \cellcolor{neg5}66.35 $\pm$ 0.72
& \cellcolor{pos5}45.91 $\pm$ 2.56
& \cellcolor{pos7}47.70 $\pm$ 7.79
& \cellcolor{pos1}40.60 $\pm$ 0.11 \\

& EvoPrompt(GA)
& \cellcolor{pos10}50.53 $\pm$ 7.20
& \cellcolor{neg3}68.50 $\pm$ 3.60
& \cellcolor{pos5}45.24 $\pm$ 3.49
& \cellcolor{pos10}57.24 $\pm$ 4.26
& \cellcolor{pos3}42.62 $\pm$ 9.11 \\

& I
& \cellcolor{pos7}42.65 $\pm$ 0.29
& \cellcolor{neg6}65.00 $\pm$ 1.29
& \cellcolor{pos7}48.33 $\pm$ 8.60
& \cellcolor{pos8}50.79 $\pm$ 1.72
& \cellcolor{neg2}39.06 $\pm$ 2.89 \\

& I+R
& \cellcolor{pos10}62.66 $\pm$ 4.68
& \cellcolor{pos4}\textbf{74.47 $\pm$ 2.97}
& \cellcolor{pos10}62.04 $\pm$ 6.18
& \cellcolor{pos10}71.70 $\pm$ 6.74
& \cellcolor{pos10}63.87 $\pm$ 6.00 \\

& I+R+E
& \cellcolor{pos10}60.23 $\pm$ 12.69
& \cellcolor{neg1}70.41 $\pm$ 3.83
& \cellcolor{pos10}74.30 $\pm$ 6.67
& \cellcolor{pos10}63.79 $\pm$ 7.58
& \cellcolor{pos10}60.62 $\pm$ 15.46 \\

\midrule


\multirow{12}{*}{Stay at home orders}

& Base
& 58.47 $\pm$ 0.32
& 75.67 $\pm$ 0.83
& 64.88 $\pm$ 3.53
& 72.80 $\pm$ 0.49
& 76.36 $\pm$ 0.50 \\

& Base+CoT
& \cellcolor{pos10}74.52 $\pm$ 1.42
& \cellcolor{pos2}75.96 $\pm$ 0.66
& \cellcolor{pos8}74.83 $\pm$ 0.71
& \cellcolor{pos3}75.32 $\pm$ 0.42
& \cellcolor{pos1}76.01 $\pm$ 0.36 \\

& Base+FS
& \cellcolor{pos3}61.04 $\pm$ 1.98
& \cellcolor{pos1}76.26 $\pm$ 1.68
& \cellcolor{pos8}74.30 $\pm$ 1.20
& \cellcolor{pos6}79.29 $\pm$ 0.50
& \cellcolor{pos5}80.53 $\pm$ 0.00 \\

& Base+CoT+FS
& \cellcolor{pos10}74.34 $\pm$ 4.20
& \cellcolor{pos2}75.85 $\pm$ 1.46
& \cellcolor{pos10}79.54 $\pm$ 1.22
& \cellcolor{pos8}\textbf{81.33 $\pm$ 0.55}
& \cellcolor{pos5}80.14 $\pm$ 0.04 \\
& Base+CoS
& \cellcolor{neg10}43.38 $\pm$ 3.39
& \cellcolor{neg8}65.35 $\pm$ 2.80
& \cellcolor{pos5}69.61 $\pm$ 3.33
& \cellcolor{neg5}68.27 $\pm$ 2.37
& \cellcolor{neg6}69.61 $\pm$ 0.69 \\

& OPRO
& \cellcolor{pos8}69.24 $\pm$ 4.80
& \cellcolor{neg3}72.69 $\pm$ 4.86
& \cellcolor{pos3}67.41 $\pm$ 0.34
& \cellcolor{neg5}68.73 $\pm$ 3.41
& \cellcolor{neg2}74.81 $\pm$ 1.47 \\

& CAPO
& \cellcolor{pos9}71.90 $\pm$ 3.91
& \cellcolor{neg1}75.20 $\pm$ 1.73
& \cellcolor{pos5}69.74 $\pm$ 0.65
& \cellcolor{neg2}71.38 $\pm$ 3.11
& \cellcolor{neg3}73.80 $\pm$ 4.01 \\

& EvoPrompt(DE)
& \cellcolor{pos8}69.04 $\pm$ 0.73
& \cellcolor{neg2}74.57 $\pm$ 0.04
& \cellcolor{pos5}69.30 $\pm$ 4.27
& \cellcolor{pos2}73.92 $\pm$ 1.36
& \cellcolor{neg4}72.59 $\pm$ 1.10 \\

& EvoPrompt(GA)
& \cellcolor{pos8}69.12 $\pm$ 1.15
& \cellcolor{neg3}73.30 $\pm$ 2.69
& \cellcolor{pos6}70.36 $\pm$ 2.27
& \cellcolor{neg2}71.55 $\pm$ 4.65
& \cellcolor{neg4}72.67 $\pm$ 3.28 \\

& I
& \cellcolor{pos6}64.83 $\pm$ 2.97
& \cellcolor{neg2}74.15 $\pm$ 1.88
& \cellcolor{pos7}73.30 $\pm$ 4.56
& \cellcolor{pos5}77.73 $\pm$ 1.64
& \cellcolor{neg3}73.67 $\pm$ 2.45 \\

& I+R
& \cellcolor{pos9}72.93 $\pm$ 3.22
& \cellcolor{pos1}75.99 $\pm$ 0.86
& \cellcolor{pos9}77.10 $\pm$ 3.33
& \cellcolor{pos5}76.85 $\pm$ 0.55
& \cellcolor{neg2}74.58 $\pm$ 1.38 \\

& I+R+E
& \cellcolor{pos9}73.12 $\pm$ 0.87
& \cellcolor{neg8}64.73 $\pm$ 8.46
& \cellcolor{pos8}76.85 $\pm$ 2.68
& \cellcolor{pos4}76.59 $\pm$ 3.89
& \cellcolor{pos1}76.81 $\pm$ 3.72 \\

\bottomrule
\end{tabular}
}

\caption{
Per-target results on COVID-19-Stance using Macro-F1.
I = instruction-only; I+R = instruction + reasoning; I+R+E = instruction
+ per-example reasoning. Base = original prompt; CoT = chain-of-thought
prompting; FS = few-shot prompting; CoS = Chain-of-Stance prompting.
Values are mean $\pm$ standard deviation across three runs.
Shaded cells indicate differences from the corresponding Base score:
gray denotes improvements, red denotes decreases, and darker shades indicate
larger absolute differences. The best score for each target across all
models and methods is shown in \textbf{bold}.
}
\label{tab:covid-19-stance-results-po}

\end{table*}

\section{Few-Shot Prompting Analysis}
\label{app:fewshot_analysis}

We further evaluate whether retrieved demonstrations improve the final
I+R+E prompts. We compare I+R+E without in-context demonstrations at
inference (zero-shot) with few-shot variants using $k=1$ and $k=3$
retrieved demonstrations per stance class across all four datasets and
five backbone models.

\begin{figure*}[!t]
    \centering
    \includegraphics[width=\textwidth]{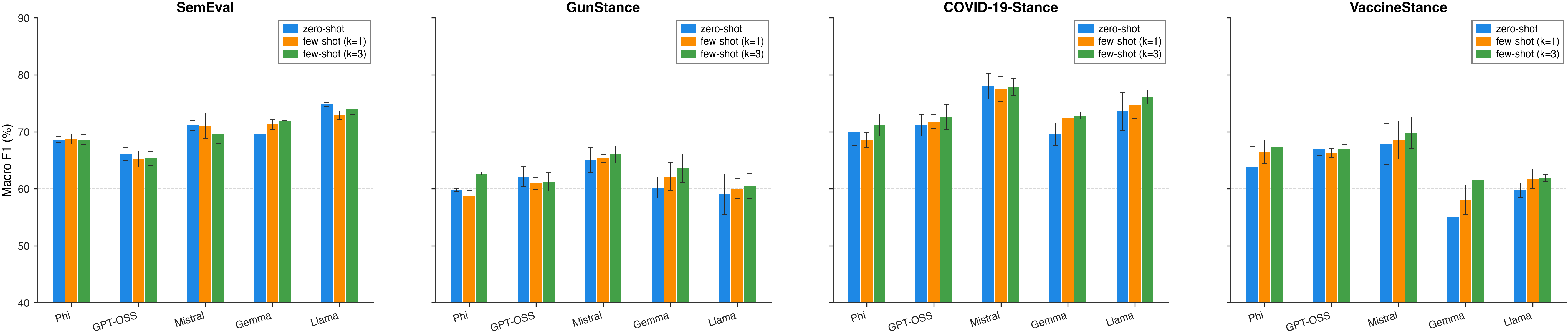}
    \caption{
    Macro-F1 comparison of zero-shot and few-shot prompting with the
    final I+R+E prompts across four benchmark datasets. Zero-shot uses
    no in-context demonstrations at inference; few-shot uses $k=1$ or
    $k=3$ retrieved demonstrations per stance class. Bars show means
    over three runs; error bars denote standard deviation.
    }
    \label{fig:ire-zeroshot-fewshot-k1-k3}
\end{figure*}

As shown in Figure~\ref{fig:ire-zeroshot-fewshot-k1-k3}, the effect of
few-shot prompting varies across models and datasets. Retrieved
demonstrations improve Macro-F1 in several settings, particularly on
GunStance and VaccineStance, but provide smaller gains or decreases in
others. Increasing $k$ from 1 to 3 also does not consistently improve
performance, indicating that additional demonstrations are not
uniformly beneficial after I+R+E optimization.

\section{External Reasoner Results}
\label{app:gpt5_reasoner}

Table~\ref{tab:gpt5_optimizer_comparison} reports the full results of
the external-reasoner experiment summarized in Table~\ref{tab:gpt5-optimizer-comparison}
of the main paper. We compare I+R+E using the classifier itself as the
reasoner (\textit{Same}) with GPT-5 as an external reasoner
(\textit{GPT-5}) across all four datasets and five backbone models.

\begin{table*}[!t]
\centering
\begingroup
\fontsize{7.2}{8.2}\selectfont
\setlength{\tabcolsep}{5pt}

\begin{tabular}{llccccc}
\toprule
\textbf{Dataset} & \textbf{Reasoner}
& \textbf{Phi}
& \textbf{GPT-OSS}
& \textbf{Mistral}
& \textbf{Gemma}
& \textbf{Llama} \\
\midrule

\multirow{2}{*}{SemEval}
& I+R+E (Same)
& 68.63 $\pm$ 0.53
& 66.10 $\pm$ 1.14
& 71.12 $\pm$ 0.85
& 69.68 $\pm$ 1.14
& 74.78 $\pm$ 0.38 \\

& I+R+E (GPT-5)
& \cellcolor{pos1}69.11 $\pm$ 0.26
& \cellcolor{pos1}66.35 $\pm$ 1.17
& \cellcolor{neg3}65.30 $\pm$ 2.46
& \cellcolor{pos3}71.91 $\pm$ 1.21
& \cellcolor{pos2}76.46 $\pm$ 0.27 \\
\midrule

\multirow{2}{*}{GunStance}
& I+R+E (Same)
& 59.74 $\pm$ 0.26
& 62.12 $\pm$ 1.76
& 65.01 $\pm$ 2.18
& 60.21 $\pm$ 1.84
& 59.03 $\pm$ 3.58 \\

& I+R+E (GPT-5)
& \cellcolor{pos2}61.61 $\pm$ 1.40
& \cellcolor{neg2}60.56 $\pm$ 0.04
& \cellcolor{neg1}64.14 $\pm$ 1.08
& \cellcolor{pos3}62.99 $\pm$ 1.34
& \cellcolor{pos8}68.45 $\pm$ 1.75 \\
\midrule

\multirow{2}{*}{COVID-19-Stance}
& I+R+E (Same)
& 69.97 $\pm$ 2.43
& 71.14 $\pm$ 1.89
& 78.03 $\pm$ 2.24
& 69.55 $\pm$ 1.96
& 73.58 $\pm$ 3.29 \\

& I+R+E (GPT-5)
& \cellcolor{pos1}70.07 $\pm$ 1.07
& \cellcolor{pos1}71.21 $\pm$ 0.57
& \cellcolor{pos2}79.07 $\pm$ 0.26
& \cellcolor{pos7}77.99 $\pm$ 0.17
& \cellcolor{pos8}82.81 $\pm$ 0.10 \\
\midrule

\multirow{2}{*}{VaccineStance}
& I+R+E (Same)
& 63.87 $\pm$ 3.57
& 66.99 $\pm$ 1.20
& 67.85 $\pm$ 3.59
& 55.12 $\pm$ 1.83
& 59.76 $\pm$ 1.27 \\

& I+R+E (GPT-5)
& \cellcolor{pos3}66.14 $\pm$ 1.67
& \cellcolor{pos2}68.01 $\pm$ 1.28
& \cellcolor{pos4}71.51 $\pm$ 0.58
& \cellcolor{pos8}66.76 $\pm$ 3.21
& \cellcolor{pos7}68.31 $\pm$ 1.85 \\

\bottomrule
\end{tabular}

\caption{
Macro-F1 comparison of I+R+E using the classifier itself
(\textit{Same}) or GPT-5 as an external reasoner.
Results are mean $\pm$ standard deviation across three independent runs.
Shaded GPT-5 cells indicate differences relative to the corresponding
Same setting; gray denotes improvement and red denotes degradation,
with darker shading indicating larger absolute differences.
}
\label{tab:gpt5_optimizer_comparison}

\endgroup
\end{table*}

\section{Additional Qualitative Examples}
\label{app:error_analysis}

\subsection{Qualitative Examples of Frozen-Module Ablation}
\label{app:frozen_examples}

To analyze the effect of freezing stance-definition modules during
optimization, we examine cases where the full I+R+E configuration
predicts the correct stance label but the corresponding frozen-module
variant fails. These examples illustrate cases in which freezing
individual modules is associated with errors involving implicit stance
cues and neutrality-related expressions.

The following examples present prediction flips from the frozen-module ablation analysis using Mistral3.2 on SemEval-2016.

\vspace{0.5em}
\begin{casebox}{Case 1: Favor-Definition Freezing Failure}
\small

\textbf{Dataset:} SemEval-2016 \\
\textbf{Target:} Legalization of Abortion

\vspace{2pt}

\textbf{Tweet:}
\emph{``Only one can have bodily autonomy \#SemST''}

\vspace{3pt}

{\small
\textbf{Gold:} \labeltag{correctbg}{\it favor}
\hspace{0.6cm}

\textbf{Full I+R+E:} \labeltag{correctbg}{\it favor}
\hspace{0.6cm}

\textbf{Frozen:} \labeltag{errorbg}{\it against}
}

\end{casebox}

\vspace{-0.65cm}

\begin{casebox}{Case 2: Against-Definition Freezing Failure}
\small

\textbf{Dataset:} SemEval-2016 \\
\textbf{Target:} Atheism

\vspace{2pt}

\textbf{Tweet:}
\emph{``Remember, \#God has it all worked out. \#SemST''}

\vspace{3pt}

{\small
\textbf{Gold:} \labeltag{correctbg}{\it against}
\hspace{0.3em}

\textbf{Full I+R+E:} \labeltag{correctbg}{\it against}
\hspace{0.3em}

\textbf{Frozen:} \labeltag{errorbg}{\it neutral}
}

\end{casebox}

\vspace{-0.65cm}

\begin{casebox}{Case 3: Neutral-Definition Freezing Failure}
\small

\textbf{Dataset:} SemEval-2016 \\
\textbf{Target:} Legalization of Abortion

\vspace{2pt}

\textbf{Tweet:}
\emph{``Like how can you sit there and even go through with the process \#SemST''}

\vspace{3pt}

{\small
\textbf{Gold:} \labeltag{correctbg}{\it neutral}
\hspace{0.3em}

\textbf{Full I+R+E:} \labeltag{correctbg}{\it neutral}
\hspace{0.3em}

\textbf{Frozen:} \labeltag{errorbg}{\it against}
}

\end{casebox}

\subsection{Qualitative Case Studies}
\label{sec:qualitative_case_studies}

To illustrate how the optimized prompts affect individual stance
predictions, we examine representative cases in which optimization
corrects errors made by the Base prompt. The examples highlight cases
involving implicit stance and neutrality disambiguation, where different
levels of optimization lead to different predictions.

\begin{casebox}{Case 1: Implicit Environmental Advocacy with I}
\small

\textbf{Dataset:} SemEval-2016 \\
\textbf{Target:} Climate Change is a Real Concern

\vspace{3pt}

\textbf{Tweet:}
\emph{``grow food not lawns Lawns are Ecocide in your yard. \#Patriarchy is Ecocide of your own species \#DominoEffect \#SemST''}

\vspace{3pt}

\textbf{Gold:} \labeltag{correctbg}{\it favor}
\quad
\textbf{Base:} \labeltag{errorbg}{\it against}
\quad
\textbf{I:} \labeltag{correctbg}{\it favor}
\quad
\textbf{I+R:} \labeltag{correctbg}{\it favor}
\quad
\textbf{I+R+E:} \labeltag{correctbg}{\it favor}

\vspace{3pt}

\textbf{Why optimization succeeds:}
The tweet implicitly supports environmental concerns despite its negative tone.
While the Base prompt relies on surface sentiment and predicts \emph{against},
the optimized variants correctly identify the underlying supportive stance.

\end{casebox}

\vspace{-0.05cm}

\begin{casebox}{Case 2: Neutrality Disambiguation with I+R}
\small

\textbf{Dataset:} SemEval-2016 \\
\textbf{Target:} Climate Change is a Real Concern

\vspace{3pt}

\textbf{Tweet:}
\emph{``Boni `Future Sea level rise will not be uniform \& different for coasts' \#CCFC15 \#Journey2015 \#SemST''}

\vspace{3pt}

\textbf{Gold:} \labeltag{correctbg}{\it favor}
\quad
\textbf{Base:} \labeltag{errorbg}{\it neutral}
\quad
\textbf{I:} \labeltag{errorbg}{\it neutral}
\quad
\textbf{I+R:} \labeltag{correctbg}{\it favor}
\quad
\textbf{I+R+E:} \labeltag{correctbg}{\it favor}

\vspace{3pt}

\textbf{Why optimization succeeds:}
The tweet presents scientific evidence about climate change without explicit
advocacy. While Base and I predict \emph{neutral}, the reasoning-enhanced
variants correctly identify the implicit supportive stance.

\end{casebox}

\end{document}